\documentclass[pdflatex,sn-nature]{sn-jnl}

\usepackage{graphicx}%
\usepackage{multirow}%
\usepackage{amsmath,amssymb,amsfonts}%
\usepackage{amsthm}%
\usepackage{mathrsfs}%
\usepackage[title]{appendix}%
\usepackage{xcolor}%
\usepackage{textcomp}%
\usepackage{manyfoot}%
\usepackage{booktabs}%
\usepackage{algorithm}%
\usepackage{algorithmicx}%
\usepackage{algpseudocode}%
\usepackage{listings}%
\usepackage{tcolorbox}
\tcbuselibrary{skins,breakable}

\usepackage{tcolorbox}
\usepackage{xcolor}
\usepackage{listings}
\usepackage{courier}
\usepackage{setspace}
\usepackage{afterpage}

\tcbset{
  colback=gray!4,
  colframe=black,
  arc=2mm,
  boxrule=0.8pt,
  left=5pt,
  right=5pt,
  top=5pt,
  bottom=5pt,
  fonttitle=\bfseries,
  coltitle=white,
  toptitle=2pt,
  bottomtitle=1pt,
  title filled,
}

\lstdefinestyle{jsonstyle}{
  basicstyle=\ttfamily\footnotesize,
  backgroundcolor=\color{gray!5},
  frame=single,
  breaklines=true,
  showstringspaces=false,
  numbers=none
}

\theoremstyle{thmstyleone}%
\theoremstyle{thmstyletwo}%

\theoremstyle{thmstylethree}%

\begin{document}

\title[Article Title]{Universally Converging Representations of Matter Across Scientific Foundation Models}


\author[1]{\fnm{Sathya} \sur{Edamadaka}}\email{sathyae@mit.edu}
\equalcont{These authors contributed equally to this work.}

\author*[2]{\fnm{Soojung} \sur{Yang}}\email{soojungy@mit.edu}
\equalcont{These authors contributed equally to this work.}


\author[1,3]{\fnm{Ju} \sur{Li}}\email{liju@mit.edu}

\author*[1]{\fnm{Rafael} \sur{Gómez-Bombarelli}}\email{rafagb@mit.edu}

\affil[1]{\orgdiv{Department of Materials Science and Engineering}, \orgname{Massachusetts Institute of Technology}, \orgaddress{\street{77 Massachusetts Avenue}, \city{Cambridge}, \postcode{02139}, \state{MA}, \country{USA}}}

\affil[2]{\orgdiv{Computational \& Systems 
Biology}, \orgname{Massachusetts Institute of Technology}, \orgaddress{\street{77 Massachusetts Avenue}, \city{Cambridge}, \postcode{02139}, \state{MA}, \country{USA}}}

\affil[3]{\orgdiv{Department of Nuclear Science and Engineering}, \orgname{Massachusetts Institute of Technology}, \orgaddress{\street{77 Massachusetts Avenue}, \city{Cambridge}, \postcode{02139}, \state{MA}, \country{USA}}}


\abstract{
Machine learning models of vastly different modalities and architectures are being trained to predict the behavior of molecules, materials, and proteins. However, it remains unclear whether they learn similar internal representations of matter. Understanding their latent structure is essential for building scientific foundation models that generalize reliably beyond their training domains. Although representational convergence has been observed in language and vision, its counterpart in the sciences has not been systematically explored. Here, we show that representations learned by nearly sixty scientific models, spanning string-, graph-, 3D atomistic, and protein-based modalities, are highly aligned across a wide range of chemical systems. Models trained on different datasets have highly similar representations of small molecules, and machine learning interatomic potentials converge in representation space as they improve in performance, suggesting that foundation models learn a common underlying representation of physical reality. We then show two distinct regimes of scientific models: on inputs similar to those seen during training, high-performing models align closely and weak models diverge into local sub-optima in representation space; on vastly different structures from those seen during training, nearly all models collapse onto a low-information representation, indicating that today's models remain limited by training data and inductive bias and do not yet encode truly universal structure. Our findings establish representational alignment as a quantitative benchmark for foundation-level generality in scientific models. More broadly, our work can track the emergence of universal representations of matter as models scale, and for selecting and distilling models whose learned representations transfer best across modalities, domains of matter, and scientific tasks. 
}

\maketitle

\section{Introduction}\label{sec1}

Artificial intelligence has undergone a paradigm shift from bespoke, task-specific models to general-purpose ``foundation models" \cite{foundationmodelref}. These models are pre-trained on vast and diverse datasets. As a result, they can perform a wide range of downstream tasks that they were never explicitly taught to solve. The key to this emergent capability is their representational power: foundation models learn compact, latent representations of each input, enabling strong performance on data far beyond the examples seen during training.  

The success of foundation models in language and vision has motivated a parallel effort in the sciences. 
Across chemistry and biology, early approaches relied on hand-crafted, one-dimensional features. Atomic number, hybridization, and other chemically derived descriptors were calculated for molecules \cite{rdkit}, and one-hot amino acid vectors for proteins \cite{elabd2020amino}. Two dimensional input modalities also became widely used, including SMILES strings for molecules \cite{smiles}, crystal graphs for materials \cite{cgcnn}, and amino acid string sequences for proteins \cite{qian1988predicting}. Self-supervised encoders trained on these input modalities learned even richer, higher-dimensional representations than hand-crafted features alone, achieving improved performance on downstream chemical property prediction tasks \cite{molformer2022, Pengmei2025geom2vec}. For proteins, large self-supervised sequence models like ESM2 showed that learned representations can support a variety of tasks, including structure prediction, design, and functional annotation tasks \cite{esm2}.

Beyond 1D and 2D modalities, many models now learn directly from 3D atomic coordinates. Leveraging increasingly large and chemically diverse simulated datasets, large-scale, supervised models have been trained to predict interatomic forces and energies in molecular and materials systems \cite{batatia2023mace, orbv3}. These models, often called machine learning interatomic potentials (MLIPs), were originally developed for molecular simulation, but are now being repurposed for a growing range of scientific downstream tasks \cite{Li25UniversalMLIPShines}.
For instance, MLIP representations were shown to capture local protein geometries and chemical environments well enough to yield accurate NMR chemical shift prediction \cite{bojan2025representinglocalproteinenvironments}, and universal MLIPs can be utilized for out-of-domain electronic structure and excited state predictions \cite{KimParkLi}. Notably, models using MLIP representations produced more physically consistent predictions than task-specific baselines \cite{KimParkLi}, showing how force- and energy-based training can effectively learn fundamental structure–energy relationships and  physically ground these models.

Despite their diverse input modalities, architectures, and training data domains, these scientific foundation models ultimately seek to learn the same underlying physical principles. Therefore, just as scientists reason across chemistry, materials science, and biology using a unified set of laws, there may exist a single statistical model that captures the joint distribution over materials and their properties. Such a model would encode a unified representation of matter across scientific domains, revealing a common structure in the behavior of molecules, materials, and proteins. Recent analogous efforts outside scientific domains provide inspiring evidence for this. As vision and language models improved in performance, their learned representations grew increasingly similar, hypothetically converging to a true understanding of reality \cite{huh2024prh}. Separately, representations of text from language models with vastly different architectures, training data, and sizes were directly translated without any paired data labels, showing the similarity of the underlying information they learned \cite{jha2025harnessinguniversalgeometryembeddings}. Therefore, given such substantial evidence in other domains, it is natural to ask if scientific foundation models are converging towards a universal representation of matter.

In this work, we find that scientific foundation models of different modalities, training tasks, and architectures have significantly aligned latent representations. We then find that as models improve in performance, their representations converge, suggesting that foundation models learn a common underlying representation of physical reality. We then establish a dynamic benchmark for foundation-level generality by probing representations of in-distribution structures already seen by models and out-of-distribution, unseen structures. Lastly, we suggest several lessons for future scientific model development that arise from our analysis. 

Our study covers 59 models across multiple input modalities (SMILES/SELFIES-string encodings of molecules, 3D atomistic coordinates, protein sequences, protein structures, and natural language), architectures (equivariant and non-equivariant MLIPs, conservative and direct prediction models), and training domains (molecules, materials, and proteins). We compare representations of matter from five datasets (molecules from QM9 \cite{moleculenet} and OMol25 \cite{levine2025openmolecules2025omol25}, materials from OMat24 \cite{barroso_omat24} and sAlex \cite{sAlex}, and proteins from RCSB \cite{rcsb}). Concretely, we generate representations by inputting structures from these datasets into each model and saving numerical embeddings from their last hidden layer. We measure representational alignment, or how similar two models' latent spaces are, with four vastly different metrics that operate directly on model embeddings.

\section{Results}
\subsection{Representations of matter in scientific foundation models are converging}\label{sec2}

\afterpage{%
\clearpage
\begingroup
\addtolength{\footskip}{3cm} 
\begin{figure}[p]
    \centering
    \makebox[\textwidth]{\includegraphics[width=1.3\linewidth]{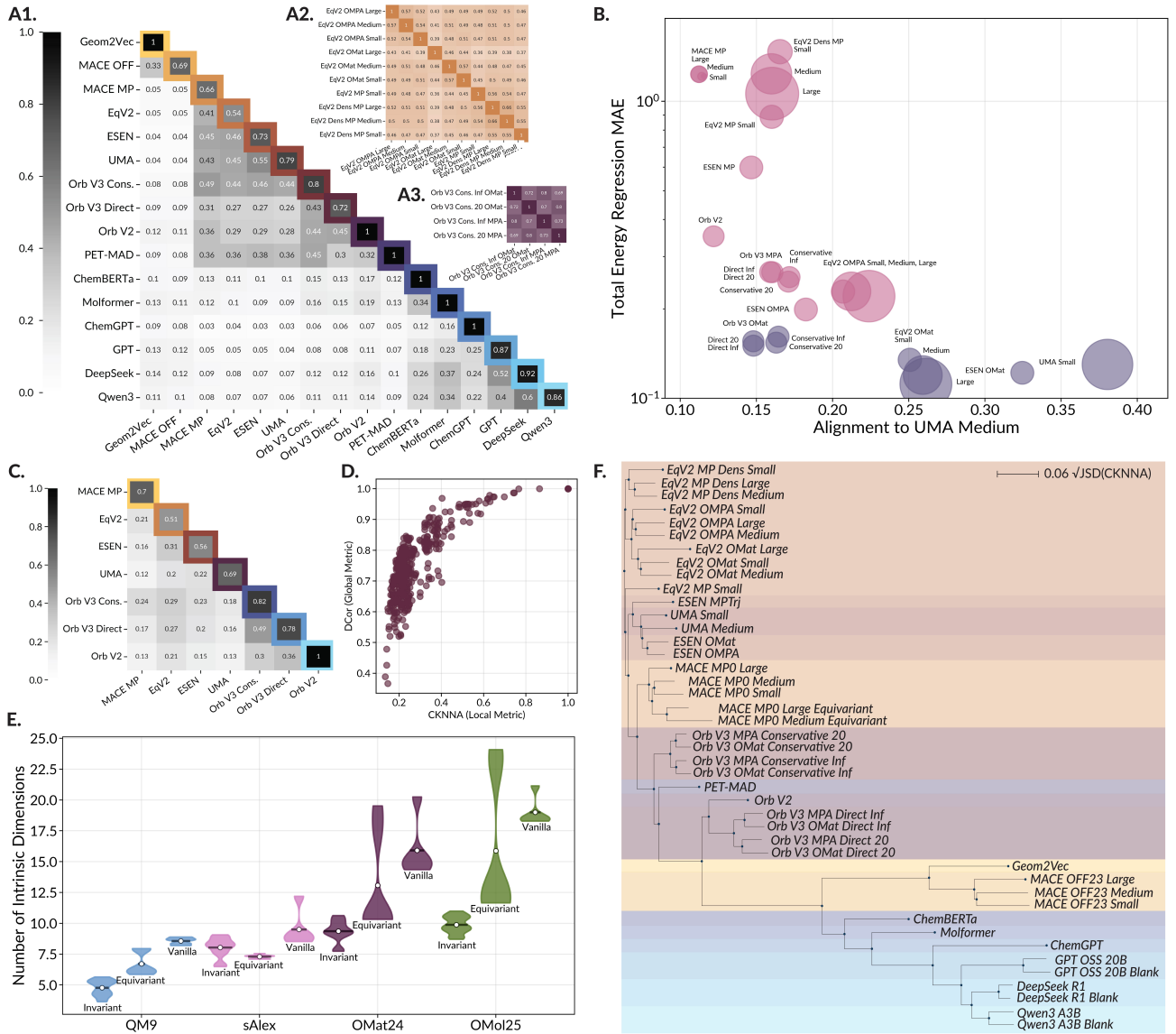}}
    \makebox[\textwidth]{%
    \begin{minipage}{1.3\linewidth}
    \caption{\small Evidence for the convergence of scientific foundation model representations. A1 shows representational alignment between model architectures as measured by Centered Kernel Nearest-Neighbor Alignment (CKNNA) on structures from QM9 \cite{moleculenet}. It is a condensed version of Fig. \ref{fig:qm9_full_cknna}, as described in the Methods section, where each cell represents the average of several rows and columns. Some elements along the block diagonal are not 1 because we averaged several configurations (e.g., size, training data) of the same model architecture together for easier visualization. The full block along Fig. \ref{fig:qm9_full_cknna}'s diagonal for Equiformer V2 models and Orb V3 models is shown in A2 and A3. The off-diagonal cells between Molformer (text-based) and Orb V2 and Orb V3 (3D atom coordinate-based) models show significant cross-modality alignment.
    B. shows the convergence of scientific model representations—namely, as models improve in energy regression mean absolute error (MAE) performance, their representational similarity to the best performing model (UMA Medium) increases. Each point represents a single model, and its size is proportional to the size of the model. We observe the same effect within each cluster of models within the same family. 
    C shows a similar condensed CKNNA correlation matrix as A1 but for OMat24. 
    D shows how CKNNA, a local metric, is highly correlated with DCor, a purely global metric, showing that our analysis is applicable to beyond only local environments in representation space. 
    E shows distributions of the number of intrinsic dimensions ($I_d$) of model embeddings for different datasets through violin plots. We observe larger $I_d$ for structures with more compositional complexity, increased $I_d$ when including non-equilibrium structures (OMat24 vs. sAlex). Models of vastly different architectures largely have similar intrinsic dimensionalities. In addition, models that output invariant embeddings have lower intrinsic dimensionalities than equivariant model and vanilla (neither equivariant nor invariant) model representations.
    F shows an evolutionary tree of scientific foundation models, using CKNNA representational similarity as a proxy of evolutionary distance. Models with more similar representation spaces are closer together. The tree groups models of the same dataset, equivariance, and architecture together, in that order. }
    \label{fig:1}
    \end{minipage}%
    }
\end{figure}
\clearpage
\endgroup
}

To explore whether scientific foundation models learn similar representations, we first examine embeddings from models trained on vastly different input modalities, from string-based encodings and two-dimensional graphs of molecules to three-dimensional atomic coordinates of materials. Despite the vastly different training datasets and input modalities between models, we observe significant representational alignment. This alignment grows stronger as model performance increases, indicating that models are converging to a shared representation of underlying physical principles. We show that our findings are consistent across several measures of representational similarity. We also find that model embeddings have remarkably similar intrinsic dimensionalities, implying a statistical convergence in the complexity of their latent spaces. Finally, we build an evolutionary tree of scientific foundation models, highlighting the roles of architecture and training data in shaping representational similarity.  

\subsubsection{Models are strongly aligned within and across modalities}\label{sec2a}

We first assess representational alignment between models trained on vastly different input data modalities. We generate embeddings using structures from the multi-modal QM9 dataset of small molecules, containing SMILES strings, SELFIES strings \cite{selfies}, and 3D atomic coordinates \cite{moleculenet}. The condensed alignment matrix is shown in Fig. \ref{fig:1}A, which is generated by averaging the full representation similarity matrix (Fig. \ref{fig:qm9_full_cknna}) across models of the same architecture (Fig. \ref{fig:1}A2 and \ref{fig:1}A3), as further detailed in Methods. We use the Centered Kernel Nearest-Neighbor Alignment (CKNNA) metric to measure representational similarity, which was first proposed to analyze cross-modality alignment of language and vision models \cite{huh2024prh}.

We find that models trained on the same data modality, such as MLIPs that take in 3D atom coordinates, are strongly aligned, as seen by the dark, off-diagonal triangle in Fig. \ref{fig:1}A1. This trend is also apparent for models that operate on string encodings of molecules, as in the bottom right off-diagonal triangle in Fig. \ref{fig:1}A1. 
Crucially, we observe alignment between string-based models and atomistic MLIPs, with particularly strong alignment between SMILES-based models and Orb architectures. Although their CKNNA values may seem low compared to those for within-modality alignment, they exceed the highest representation alignment values found between language and vision foundation models \cite{huh2024prh}. 
Although string encodings do not capture conformational geometry of the molecule, they contain information that is effectively equivalent to molecular graphs. Because the conformers in QM9 have minimal structural variance from their lowest-energy conformers, these molecular graphs capture most of the informational content relevant to the QM9 dataset, which explains the nontrivial alignment between string-based and 3D coordinate-based models. 
Surprisingly, large natural language models (LLMs) align strongly with string encoding-based materials models when provided with SMILES strings, and show similar alignment scores with MLIPs as other SMILES-based models. 

Protein models exhibit even stronger cross-modality alignment. As shown in Fig. \ref{fig:RCSB_cknna}, representations from protein sequence models and protein structure models align nearly twice as strongly as the best cases for small molecules. This heightened alignment is consistent with evidence that large protein sequence models implicitly learn folding constraints and structural regularities \cite{esm2, esm3, sequencetostructure}, bringing their latent spaces inherently closer to those of structure-based models.


\subsubsection{Representations are converging as model performance improves}\label{sec3}
Vastly different scientific models are not just highly aligned; they are converging to a universal representation of matter. We show in Fig. \ref{fig:1}B that as models improve at predicting the total energy of inputted materials, which is also their training task, they grow more aligned with the best-performing model. This trend is also visible within each model family, where, for instance, the EqV2 OMPA and EqV2 OMat models increase in size and performance, their representations converge further. We use embeddings of materials from OMat24 to show this (condensed model alignment matrix are shown in Fig. \ref{fig:1}C), as all models included in our work that could predict energies of structures were materials models. Further details on how we calculated energy regression MAE are available in the Methods section. 

\subsubsection{Local and global alignments follow similar trends}
As CKNNA is highly sensitive to the structure of local neighborhoods in representation space, it is possible for two models to share similar global geometry while exhibiting minimal CKNNA, or vice versa. To ensure that our conclusions are not artifacts of local behavior, we characterize global representational similarity using two additional metrics. Distance correlation (dCor) is a purely global metric that generalizes the Pearson correlation coefficient to non-linear, high-dimensional manifolds \cite{dcor}, and intrinsic dimensionality ($I_d$) estimates the smallest number of variables needed to describe a model's embedding space \cite{basile2025intrinsicdimensioncorrelationuncovering}. 

We find that CKNNA, a local metric, is highly correlated with dCor, a global metric, as shown in Fig. \ref{fig:1}D, indicating that local and global alignment capture consistent structure in learned representations of small molecules. Moreover, CKNNA remains stable as its degree of locality is increased, as shown in Fig. \ref{fig:si:cknna_vs_k}. 

\subsubsection{Consistent $I_d$ across models suggests statistical convergence of representations}
We next examine the intrinsic dimensionality, $I_d$, of model embeddings, which estimates the complexity of each model's latent space. As shown in Fig. \ref{fig:1}E, the distributions of $I_d$ vary between datasets, but are remarkably consistent within each dataset. QM9 representations have relatively low intrinsic dimensionality ($I_d\sim5$), whereas OMat24 ($I_d\sim10$), sAlex ($I_d\sim8$), and OMol25 ($I_d\sim10$) require more dimensions for accurate representations. This likely reflects differences in the diversity of chemical environments and conformations sampled in each dataset. Even though QM9 and OMol25 both contain predominantly organic molecules, QM9  consists of near-equilibrium, low-energy conformers of very small molecules, whereas OMol25 contains more non-equilibrium and higher-energy conformers of much larger molecular structures, therefore spanning a broader variety of chemical environments. The same is true for sAlex and OMat24, respectively.  

Crucially, models of different architectures, trained on different data, still produce $I_d$ values within a relatively narrow band across four different datasets (Fig. \ref{fig:1}E), suggesting that representations of matter from different architectures share a universal, relatively low-dimensional structure. This is especially apparent for invariant embeddings, extracted from architectures that treat rotated structures identically to their non-rotated counterparts. Equivariant models propagate additional rotational information to accurately output how properties like inter-atomic forces transform under rotations of the inputted structures, and as a result, they have a higher $I_d$ than invariant embeddings. Vanilla models, with neither of these inductive biases baked into their model architecture, have consistently higher $I_d$.

\subsubsection{Evolutionary tree of models reveals factors that shape representation space }\label{sec2b}
To further visualize how CKNNA captures model similarity, we constructed an evolutionary tree of scientific foundation models using CKNNA-derived distances in Fig. \ref{fig:1}f (more details on how the tree was constructed can be found in Section \ref{SI:phylogenetic}). In this tree, the closer two models are to each other, the more similar their representation spaces are. Purely from CKNNA correlation values, the tree groups expected clusters together by architecture and training dataset. However, two architectures that take in 3D atomic coordinates are more closely clustered together with string-based models in the bottom right of the figure than other MLIPs. In particular, MACE trained on the small molecule (OFF) dataset diverges from nearly all other MLIP models that are trained on materials data, including MACE-MP with a very similar model architecture. These patterns highlight the stronger influence of training data on a model's representation space than its architecture. 

This clear impact of a model's training dataset highlights how differently it can represent structures based on what it was trained on. Separately, we know that scientific foundation models claim to generalize well to structures far outside of its training data distribution. Therefore, we turn our attention to using representational similarity to measure if a model is ``foundational." We then study how model embeddings change for inputs that are inside versus outside their training data distributions. 

\subsection{Diagnosing foundational status through alignment-performance analysis}\label{sec4}

\begin{figure}
    \centering
    \includegraphics[width=\linewidth]{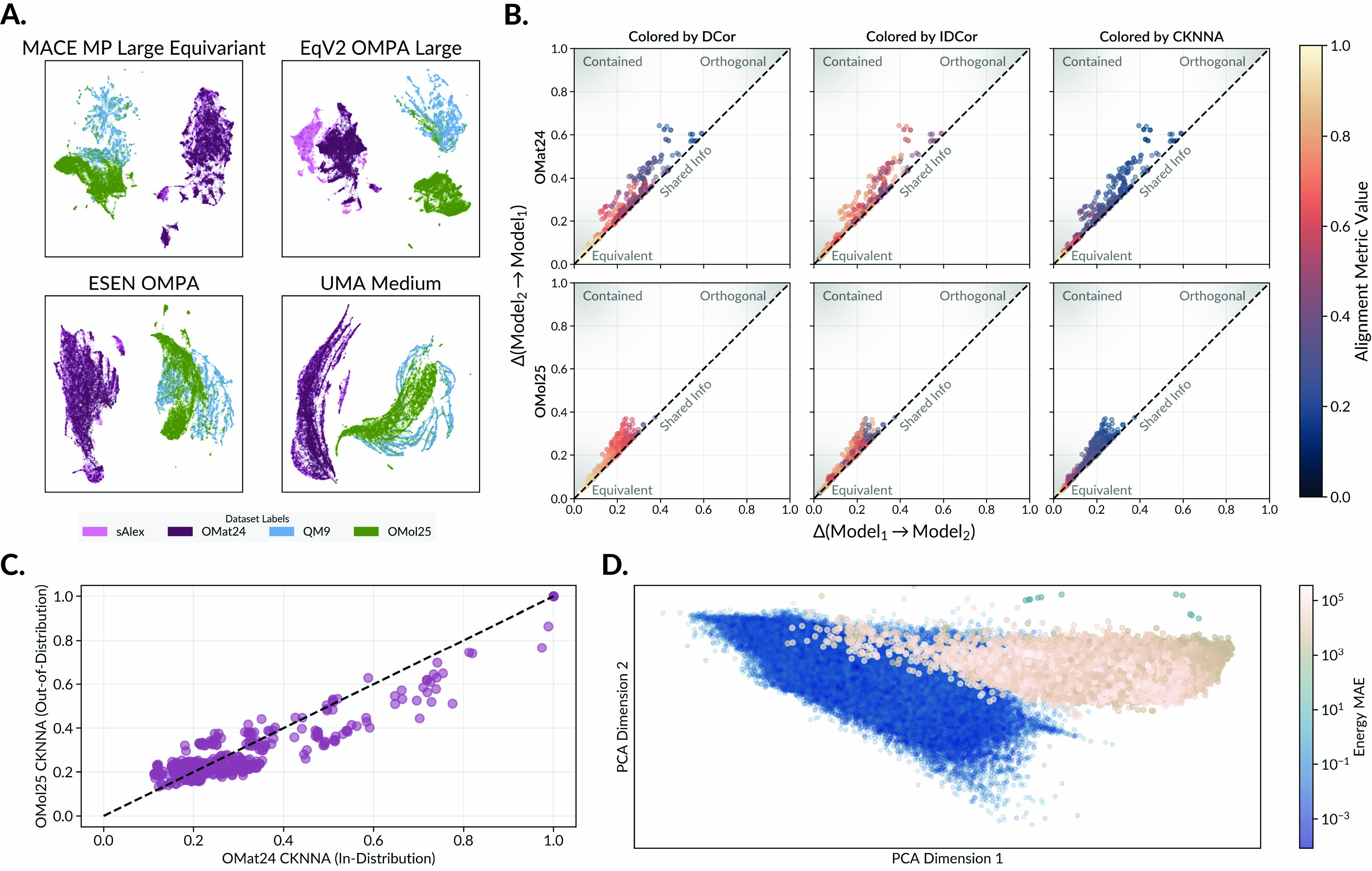}
    \caption{A shows selected model representations for 15,000 structures from each sAlex, OMat24, QM9, and OMol25, visualized in two dimensions with UMAP (the same plot for all models is shown in Fig. \ref{fig:umap_all}). Materials (OMat24 and OMol25) embeddings are largely overlapping. Molecule (QM9 and OMol25) embeddings occupy a completely different part of embedding space, falling far outside the distribution of materials embeddings.
    B shows information imbalance (II) for model embeddings of OMat24 (top row) and OMol25 (bottom row). Embeddings of in-distribution, OMat24 structures are significantly more spread out towards the top right, indicating that models represent different information. Embeddings of out-of-distribution, OMol25 structures are significantly more clustered towards the bottom left, indicating that models all represent nearly identical information. Each column includes the same information imbalance plots but colored with a different representational similarity metric, showing how CKNNA (local) and DCor (global) align very well with information imbalance, while $I_d$Cor \cite{basile2025intrinsicdimensioncorrelationuncovering}, which quantifies the correlation between different models' intrinsic dimensionality, agrees less.
    C shows the relationship between the local alignment between models for in-distribution structures (OMat24) versus out-of-distribution structures (OMol25). Although alignment in-distribution is higher locally for in-distribution structure embeddings, it is higher globally (as per information imbalance and the aligned DCors similarities (B bottom left) out-of-distribution structure embeddings.
    D shows that larger training task (energy prediction) error correlates with structures being out-of-distribution. Each point is an embedding of a structure from OMat24 or OMol25 by Orb V3 Conservative Inf OMat, and is colored by energy prediction error. The left, bluer (lower-error) cluster are OMat24 structure embeddings, and the right, whiter (higher-error) cluster that extends outside the cluster of in-distribution embeddings are OMol25 structure embeddings. }
    \label{fig:id_vs_ood}
\end{figure}

A foundation model is characterized by generalizing well on downstream tasks in a domain of interest, often one outside their training data distribution. However, performance alone cannot distinguish between a truly generalizable model and an overfit one. We characterize whether or not models have reached foundational status for a particular type of matter via their representational similarity to other high-performing models. Specifically, we find that a model which is both high-performing and well-generalized will show strong representational alignment with other well-performing models. On the other hand, an overfit high-performing model that sits in a local optimum will show weak alignment with other well-performing models.
We also expect foundation models to be more aligned with other well-performing models for in-distribution inputs than they are for out-of-distribution inputs \cite{huh2024prh}. 

We show evidence for both of these points by illustrating how in-distribution and out-of-distribution data can affect representations. We designate materials from OMat24 to be in-distribution, as most models that take in 3D atomic coordinates were trained or fine-tuned on structures from that dataset, and consider molecular structures from OMol25 to be out-of-distribution. Visual evidence of this is shown in Fig. \ref{fig:id_vs_ood}A, which visualizes how embeddings of OMol25 lie far outside the cluster of in-distribution materials structures across several models. We also quantitatively show that these molecules are out-of-distribution in Fig. \ref{fig:id_vs_ood}D, where we show embeddings of 150,000 structures each from OMat24 and OMol25 by an Orb V3 model. Each point is colored by the model's energy prediction error. Without any knowledge of the labels of each structure, we recover the cluster of OMat24 embeddings in blue (lower error) and the cluster of OMol25 embeddings in white (higher error). 

\subsubsection{Representational alignment of highly performing models}

\begin{figure}
    \centering
    \includegraphics[width=0.75\linewidth]{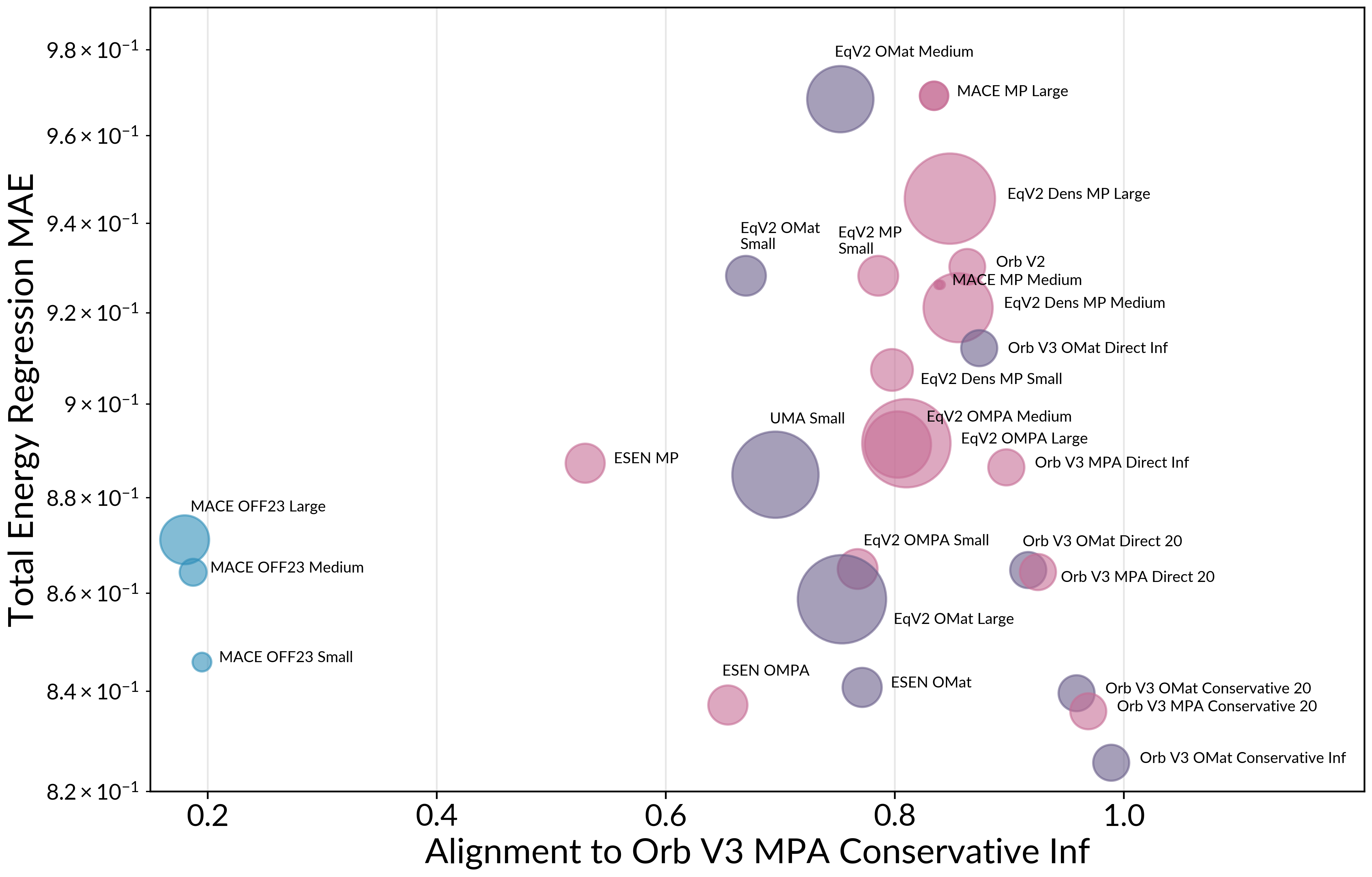}
    \caption{Scientific model representations of 1,000 structures from QM9 converge with increasing performance. Namely, as models decrease in energy regression MAE of small molecules, their representational similarity to the best performing model (Orb V3 Conservative Inf MP) increases. Each point represents a single model, and its size is proportional to the size of the model. Instead of a local metric like CKNNA to measure alignment, we use a global metric called dCor, defined in Section \ref{si:dcor}. This is because small molecules from QM9 are out-of-distribution, and local neighborhoods in representation space of out-of-distribution inputs are not guaranteed to be as meaningful as for in-distribution inputs like OMat24.}
    \label{fig:qm9_convergence}
\end{figure}

Representations of in-distribution materials show a trend of representational convergence to UMA Medium in Fig. \ref{fig:1}B. UMA therefore emerges as one of the most foundational models for materials. A similar trend appears with embeddings of structures from QM9, where models converge to an Orb V3 Conservative model, suggesting its foundational status in the chemical space of small molecules (Fig. \ref{fig:qm9_convergence}). The notable exception is the MACE-OFF model in the left of the figure; although it achieves strong performance on QM9, it shows weak alignment to the Orb V3 family, UMA, and other high-performing models. This pattern implies that MACE-OFF occupies a local optimum in representation space, whereas Orb V3 Conservative is located near a global optimum that other models consistently approach. Orb V3 substantially outperforms MACE-OFF on the GMTKN55 \cite{GMTKN55} benchmark, which is a broader and more chemically diverse molecular dataset than QM9, consistent with the interpretation that MACE-OFF’s representations are less transferable beyond the specific chemical space of QM9 \cite{Rowan}.  

\subsubsection{Weakly performing models settle into representational sub-optima} 

Representational alignment also offers diagnostic insight into how scientific models fail. We observe two distinct failure regimes. The first consists of multiple local sub-optima, where poorly performing models are weakly aligned with one another, indicating that they occupy divergent and non-generalizable regions of representation space. The second consists of a systematic lack of information, where poorly performing models remain highly aligned, suggesting that they share a common but incomplete representation missing key information critical to the domain of interest. We find that weakly performing models fall into the first failure regime on in-distribution structures but transition to the second regime when evaluated on out-of-distribution inputs. 

To evaluate the information that each model captures in its representation space, we use the information imbalance metric (a full definition is provided in section \ref{si:ii}). Unlike measures of latent space complexity ($I_d$) or global dependence (dCor), information imbalance is an asymmetric measure, explicitly quantifying which representation contains more information than another. 

On in-distribution data, weakly performing models (darker points in Fig. \ref{fig:id_vs_ood}) are weakly aligned and learn nearly orthogonal information. This dispersion indicates the presence of many local sub-optima, showing that models achieve high accuracy during training by forming idiosyncratic representations that do not generalize even to other models trained on the same domain. In this regime, the dominant principle of representation space structure is the training dataset itself. Models trained on the same dataset, despite having vastly different architectures, are consistently more aligned with one another than models sharing an architecture, but trained on different data. For instance, eSEN  models \cite{fu_esen} trained on OMat24 align more closely with EqV2 models \cite{eqv2} trained on OMat24 than they do with eSEN models trained on MPTraj (Fig. \ref{fig:OMat24_cknna}). This demonstrates that within distribution, training data has a stronger influence than architectural inductive biases on learned representations.

Out-of-distribution behavior shows the exact opposite pattern. On OMol25, almost all models fall into the second failure regime, performing poorly yet learning very similar information (Fig. \ref{fig:id_vs_ood}B, bottom row). In this case, models cluster by architecture rather than training dataset (Fig. \ref{fig:OMol25_cknna}), indicating that architectural inductive biases dominate for out-of-distribution inputs. These embeddings collapse towards architecture-specific manifolds that miss key features needed to accurately represent complex, large structures from OMol25. This is visible by comparing how many more pairs of models in the bottom row of Fig. \ref{fig:id_vs_ood}B are compressed towards the bottom left of the plot than in the top row of Fig. \ref{fig:id_vs_ood}B, thus learning more identical information to each other than for in-distribution inputs.  

Together, these contrasting failure regimes reveal that scientific model latent spaces are data-dominated in-distribution and architecture-dominated out-of-distribution. This also explains why weak models diverge into disparate suboptima in representation space or represent nearly identical, yet incomplete, information about out-of-distribution inputs. These results highlight the importance of dataset diversity for building scientific foundation models that generalize well beyond their training domains.

\section{Discussion}

Our representational analysis is a powerful tool for comparing scientific models. Therefore, we now provide several perspectives that can guide the development of scientific foundation models and help select the right model for downstream tasks. 

\subsection{Materials models are data-governed and not yet foundational}


We find that current materials models have representations that are shaped by their limited training data, and have not yet achieved foundational generality across scientific domains. As shown by visualizing information imbalance in representations of small molecules from QM9 (Fig. \ref{fig:qm9_ii_by_aspect}), models trained on the same dataset almost always learn identical information, whereas models trained on different datasets frequently learn orthogonal, non-overlapping representations. Even within the same modality, some models diverge and learn orthogonal information, indicating that existing materials training datasets do not impose a sufficiently strong statistical signal to unify these models' representation spaces. Our interpretation that models are in a data-limited regime is supported by information imbalance analysis from Section \ref{sec4}. Representations of materials structures are dominated by model training datasets, and representations of molecular structures collapse to architecture-specific manifolds that are missing chemical information. 

To reach foundational status, materials models will require substantially more diverse training data, spanning equilibrium (sAlex \cite{sAlex}) and non-equilibrium (OMat24 \cite{barroso_omat24}) regimes, and a wider range of chemical and structural environments, like those used to train Meta's UMA model \cite{wood2025umafamilyuniversalmodels}.

\subsection{Alignment emerges between models of vastly different sizes}
From our plot visualizing representational convergence with increasing performance (Fig. \ref{fig:1}B), we observe that models from the same architecture family learn similar representations (Fig. \ref{fig:1}A2, A3) even as they increase in size. This suggests that models of vastly different capacities, while differing slightly in performance, learn very similar representations. This shows that small models can mimic the expressivity of larger ones by learning similar representations, motivating the use of model distillation for downstream tasks. 

Learned representations, even from small models, can drastically accelerate generative model development. In computer vision, adding an auxiliary loss that promoted alignment with a pre-trained classifier significantly improved the training efficiency of generative models \cite{yu2024repa}. In scientific domains, a generative model sampling equilibrium conformational ensembles was trained much faster by introducing a loss regularization term for alignment with a pre-trained MLIP \cite{pinede2024REPA}. Therefore, by identifying which models learn the most transferable and universally aligned representations, our analysis provides a principled basis for selecting models that can substantially accelerate generative model training across scientific domains. 

\subsection{Representational alignment guides architectural choices}

Our representational alignment analysis reveals which architectural design choices meaningfully influence what models learn. Scientific foundation models differ widely in their inductive biases, whether they impose physical symmetries like rotational equivariance, enforce force-to-energy consistency through conservative calculations, or omit these constraints entirely to reduce computational cost. Our framework provides a principled, model-agnostic way to quantify how inductive biases shape representation spaces, allowing scientists to identify computationally inexpensive architectures that still learn expressive and transferrable representations.

To demonstrate this, we examine the recent Orb V3 family of models, which challenges the conventional view that rotational equivariance must be built into model architecture. The Orb V3 conservative variant achieves some of the strongest overall energy prediction performance (Fig. \ref{fig:omat24_maes}), yet it does not architecturally impose equivariance. Instead, it uses a lightweight regularization scheme, \textit{equigrad} \cite{orbv3}, that enforces energy quasi-invariance and force quasi-equivariance during training. Representational alignment confirms this effect: in the condensed representational similarity matrix in Fig. \ref{fig:1}A, Orb V3 conservative models align strongly with fully equivariant architectures like MACE \cite{batatia2023mace} and EqV2 \cite{eqv2}, whereas the direct Orb V3 variant without \textit{equigrad} aligns weakly. Therefore, our representational analysis recovers that appropriately structured regularization can reproduce the benefits of architectural equivariance, at substantially lower computational cost, by producing representations similar to those learned by symmetry-enforcing models. 

This result resonates with the bitter lesson in machine learning: scaling up training, rather than increasing architectural constraints or inductive biases, often yields the most general and powerful models. Although architectural equivariance is essential for simulation-focused applications of MLIPs like molecular dynamics \cite{eqv2}, our work suggests that regularization, combined with sufficient scale, can allow inexpensive architectures to approximate the representational structure of more specialized, symmetry-enforcing models.

Therefore, our representational framework not only diagnoses foundational behavior but also provides a practical guide for model selection. Scientists seeking computationally efficient models for a specific type of matter can identify smaller or cheaper architectures whose representations closely align to those of large, high-performing models. This empowers downstream tasks, like generative modeling or property prediction, to inherit expressive representational structure without the cost of training fully equivariant or conservative models.

\section{Conclusion}\label{conclusion}
We find strong evidence for the convergence of latent representations across nearly sixty scientific models. Despite differences in their input modality, architecture, and training data domain, their representations exhibit substantial alignment on small molecules. Their intrinsic dimensionalities also collapse into a narrow range, consistent with the low-dimensional structure expected of a universal representation of matter. Finally, as MLIPs improve in performance, their representations increasingly converge towards a single, unified representation.

A model approaches foundational status only when it is both high-performing and highly aligned with other performant models. Current models fall short, as their representations remain strongly data-limited. In-distribution, model representations cluster by training dataset; out-of-distribution, they collapse onto architecture-defined manifolds, revealing limited transferability across different domains of matter.

Together, these findings establish representational alignment as a powerful benchmark for diagnosing foundation-level generality in scientific models. As models continue to scale, our work can track the emergence of universal representations and guide the selection of models that best support transfer across modalities, types of matter, and scientific tasks. 

\section{Methods}\label{methods}

We sampled $N=50{,}000$ structures each from QM9, OMat24, sAlex, and OMat25 and used them to analyze all four metrics between each pair of models. All results for all datasets are available in Appendix Section C. Embeddings were extracted from each model by saving the last hidden layer output before the readout layers. Most inference on these structures was performed on a single 32 GB V100 GPU, but four 80 GB A100 GPUs were used for LLM inference. For all models that outputted node-wise embeddings, averages were taken across all nodes to produce an embedding independent of the size of each input. We list each of the models that we included in our analysis in Table \ref{tab:model_metadata}.
\begin{table*}[htbp]
\centering
\footnotesize
\renewcommand{\arraystretch}{1.1}
\setlength{\tabcolsep}{6pt}
\resizebox{\textwidth}{!}{
\begin{tabular}{l l l l l r}
\hline
\textbf{Model} & \textbf{Dataset} & \textbf{Modality} & \textbf{Embedding Type} & \textbf{DFT Level of Theory} & \textbf{\# Params} \\
\hline
Orb V2 & MPTrj, sAlex & 3D & Vanilla & PBE\_MP & 25,161,727 \\
Orb V3 OMat Direct Inf & OMat & 3D & Vanilla & PBE\_OMat24 & 25,644,479 \\
Orb V3 OMat Direct 20 & OMat & 3D & Vanilla & PBE\_OMat24 & 25,644,479 \\
Orb V3 OMat Direct 20 Randomized & OMat & 3D & Vanilla & PBE\_OMat24 & 25,644,479 \\
Orb V3 MPA Direct Inf & MPTrj, sAlex & 3D & Vanilla & PBE\_MP & 25,644,479 \\
Orb V3 MPA Direct 20 & MPTrj, sAlex & 3D & Vanilla & PBE\_MP & 25,644,479 \\
Orb V3 OMat Conservative Inf & OMat & 3D & Equivariant & PBE\_OMat24 & 25,510,582 \\
Orb V3 OMat Conservative 20 & OMat & 3D & Equivariant & PBE\_OMat24 & 25,510,582 \\
Orb V3 MPA Conservative Inf & MPTrj, sAlex & 3D & Equivariant & PBE\_MP & 25,510,582 \\
Orb V3 MPA Conservative 20 & MPTrj, sAlex & 3D & Equivariant & PBE\_MP & 25,510,582 \\
UMA Medium & OMat, OMol, OMC, ODAC, OC20 & 3D & Invariant & PBE\_OMat24 & 1,396,669,089 \\
UMA Small & OMat, OMol, OMC, ODAC, OC20 & 3D & Invariant & PBE\_OMat24 & 146,566,817 \\
ESEN MP & MPTrj & 3D & Invariant & PBE\_MP & 30,086,018 \\
ESEN OMPA & OMat, MPTrj, sAlex & 3D & Invariant & PBE\_OMat24 & 30,161,153 \\
ESEN OMat & OMat & 3D & Invariant & PBE\_OMat24 & 30,161,410 \\
EqV2 OMPA Large & OMat, MPTrj, sAlex & 3D & Invariant & PBE\_OMat24 & 153,769,868 \\
EqV2 OMPA Medium & OMat, MPTrj, sAlex & 3D & Invariant & PBE\_OMat24 & 86,589,068 \\
EqV2 OMPA Small & OMat, MPTrj, sAlex & 3D & Invariant & PBE\_OMat24 & 31,207,434 \\
EqV2 OMat Large & OMat & 3D & Invariant & PBE\_OMat24 & 153,769,868 \\
EqV2 OMat Medium & OMat & 3D & Invariant & PBE\_OMat24 & 86,589,068 \\
EqV2 OMat Small & OMat & 3D & Invariant & PBE\_OMat24 & 31,207,434 \\
EqV2 MP Small & MPTrj & 3D & Invariant & PBE\_MP & 31,207,434 \\
EqV2 Dens MP Large & MPTrj & 3D & Invariant & PBE\_MP & 160,673,293 \\
EqV2 Dens MP Medium & MPTrj & 3D & Invariant & PBE\_MP & 94,033,933 \\
EqV2 Dens MP Small & MPTrj & 3D & Invariant & PBE\_MP & 34,339,083 \\
MACE MP Large Equivariant & MPTrj & 3D & Equivariant & PBE\_MP & 15,847,440 \\
MACE MP Large & MPTrj & 3D & Invariant & PBE\_MP & 15,847,440 \\
MACE MP Medium Equivariant & MPTrj & 3D & Equivariant & PBE\_MP & 1,428,368 \\
MACE MP Medium & MPTrj & 3D & Invariant & PBE\_MP & 1,428,368 \\
MACE MP Small & MPTrj & 3D & Invariant & PBE\_MP & 3,847,696 \\
MACE OFF23 Large & SPICE & 3D & Invariant & Hybrid Functional (SPICE) & 4,707,312 \\
MACE OFF23 Medium & SPICE & 3D & Invariant & Hybrid Functional (SPICE) & 1,428,368 \\
MACE OFF23 Small & SPICE & 3D & Invariant & Hybrid Functional (SPICE) & 694,320 \\
PET-MAD & MAD, MC3D, MC2D, SHIFTML & 3D & Vanilla & PBE\_Sol & 3,258,420 \\
Geom2Vec & Denali & String / SMILES & Vanilla & None & 5,000,000 \\
Molformer & PubChem, Zinc & String / SMILES & Vanilla & None & 46,800,000 \\
ChemBERTa & PubChem & String / SMILES & Vanilla & None & 77,000,000 \\
ChemGPT & PubChem & String / SELFIES & Vanilla & None & 1,200,000,000 \\
DeepSeek R1 & Internet & String & Vanilla & None & 685,000,000,000 \\
DeepSeek R1 Blank & Internet & String & Vanilla & None & 685,000,000,000 \\
DeepSeek R1 Blank Atomistic & Internet & String & Vanilla & None & 685,000,000,000 \\
Qwen3 A3B & Internet & String & Vanilla & None & 30,000,000,000 \\
Qwen3 A3B Blank & Internet & String & Vanilla & None & 30,000,000,000 \\
Qwen3 A3B Blank Atomistic & Internet & String & Vanilla & None & 30,000,000,000 \\
GPT OSS 20B & Internet & String & Vanilla & None & 20,000,000,000 \\
GPT OSS 20B Blank & Internet & String & Vanilla & None & 20,000,000,000 \\
GPT OSS 20B Blank Atomistic& Internet & String & Vanilla & None & 20,000,000,000 \\
ESM2 Large & UniRef50 & Sequence & Vanilla & None & 650,000,000 \\ 
ESM2 Medium & UniRef50 & Sequence & Vanilla & None & 150,000,000 \\
ESM2 Small & UniRef50 & Sequence & Vanilla & None & 8,000,000 \\
ProstT5 Structure & UniRef50, PDB & Structure & Vanilla & None & 3,000,000,000 \\
ProstT5 Sequence & UniRef50, PDB & Sequence & Vanilla & None & 3,000,000,000 \\
ESMC Medium & UniRef50 & Sequence & Vanilla & None & 600,000,000 \\
ESMC Small & UniRef50 & Sequence & Vanilla & None & 300,000,000 \\
ESM3 Sequence & UniRef50, PDB & Sequence & Vanilla & None & 1,400,000,000 \\
ESM3 Structure & UniRef50, PDB & Structure & Vanilla & None & 1,400,000,000 \\
ESM IF1 & PDB & Structure & Vanilla & None & 142,000,000 \\
ProteinMPNN & PDB & Structure & Vanilla & None & 1,660,000 \\
Random Baseline & None & 3D & Vanilla & Random & -- \\
\hline
\end{tabular}
}
\caption{Summary of all models evaluated, including dataset, modality, embedding type, level of theory (if trained on DFT data), and parameter count.}
\label{tab:model_metadata}
\end{table*}

The Orb V2\cite{orbv2}, Orb V3\cite{orbv3}, UMA (OMat task selected for OMat24 and sAlex data) \cite{wood2025umafamilyuniversalmodels}, ESEN \cite{fu_esen}, Equiformer V2 \cite{eqv2}, MACE MP0 \cite{batatia2023mace}, and PET-MAD \cite{petmad} were trained on materials. UMA (OMol task selected for QM9, OMol25, and RCSB data) \cite{wood2025umafamilyuniversalmodels}, MACE OFF23 \cite{maceoff}, Geom2Vec \cite{Pengmei2025geom2vec}, Molformer \cite{molformer2022}, ChemBERTa \cite{ahmad2022chemberta2chemicalfoundationmodels}, and ChemGPT \cite{chemgpt} were trained on molecules (the latter three of which were trained on SMILES strings). ESM2 \cite{esm2}, ESM3 \cite{esm3}, ProstT5 \cite{prostt5}, ESMC \cite{esmc}, ESM Inverse Folding 1 \cite{esmif1}, and ProteinMPNN \cite{proteinmpnn} were trained on protein structure and sequences. Lastly, DeepSeek R1 (distilled onto Llama 8B) \cite{deepseek}, Qwen3 30B A3B Thinking 2507 \cite{qwen3}, and GPT OSS 20B \cite{gptoss} were all trained on natural language. Three variants of each were evaluated and shown in Figure \ref{fig:qm9_full_cknna}: one with an extended system prompt given SMILES strings, one with a minimal system prompt given SMILES strings (``... Blank") , and one with a minimal system prompt given an ASE Atoms object readout with arrays of atom 3D positions and their element types (``... Blank Atomistic"). We wanted to try a minimal system prompt to avoid introducing additional context that might dilute model attention to the wrong tokens and decrease alignment needlessly. LLM system prompts, example inputs, and example outputs are provided in Section \ref{SI:models}.

Exact definitions of each representation alignment metric are provided in appendix section A. Crucially, Fig. \ref{fig:si:cknna_vs_k} shows why we can use insights from CKNNA analysis beyond just local regimes in latent space. The block-diagonal CKNNA matrices shown in Fig. 1A. and Fig. 1C. were calculated by generating the full embedding matrices (Fig. \ref{fig:qm9_full_cknna} and Fig. \ref{fig:OMat24_cknna}, respectively) and grouping together rows (and thus columns) of the same architecture together, averaging their CKNNA values. For architectures with more than one configuration, like the eleven EqV2 \cite{eqv2} models or eight Orb V3 models, this results in a diagonal element of less than one, as it is the average of an entire subset of the original matrix along the block diagonal.

Fig. 1B, which shows the convergence of materials model representations, was constructed by finding the energy regression MAE for $1{,}000$ structures. We chose OMat24 as it was in-distribution for many of the models we included that could predict total potential energy of a structure. However, our materials models have different levels of DFT theory with different reference and atomization energies. Specifically, they either were trained on the PBE-OMat24 or the PBE-MP levels of theory. In order to compare models from both groups at the same task, we first train a linear model on the ground truth energy labels as a function of the composition, a feature vector of the counts of each element (Fig. \ref{fig:all_model_linear_compositional_dof}) and subtract the linear model's energy predictions from the true energy predictions (Fig. \ref{fig:deviations_from_linear_model_mae}). What remains is the deviation in eV from a purely compositional model's prediction of energies, which represents subtracting out the atomization energy (Fig. \ref{fig:omat24_maes}). By repeating this process for each model's energy predictions, we define energy regression MAE as the mean absolute difference of the model's deviations from its linear compositional model's predictions and the ground truth energies' deviations from its linear compositional model's predictions. 

We investigated the convergence of materials models by plotting energy regression on $1{,}000$ structures from the OMat24 dataset against alignment to the highest-performing model in Fig. 1B. We chose this dataset as it lies in-distribution for many models capable of predicting the total potential energy $E_{\text{DFT}}$ of a given structure. However, these materials models were trained on datasets with different DFT levels of theory (\textsc{OMat24} and \textsc{MP}), each corresponding to different reference and atomization energies. To approximate obtaining and subtracting out atomization energies for each level of theory, we can remove compositional bias by fitting a linear compositional model
\begin{equation}
    \hat{E}_{\text{lin}}({x}_i) = \mathbf{w}^\top \mathbf{c}_i + b,
\end{equation}
where $\mathbf{c}_i \in \mathbb{R}^M$ is a vector of elemental counts (composition) for structure ${x}_i$, and $\mathbf{w}$ and $b$ are regression parameters fit to minimize the squared error between $\hat{E}_{\text{lin}}$ and the ground truth energies $E_{\text{DFT}}({x}_i)$. 

The deviation from this linear baseline, representing the energy's deviation from atomization or structural contribution, is defined for each structure as
\begin{equation}
    \Delta E_{\text{true}}({x}_i) = E_{\text{DFT}}({x}_i) - \hat{E}_{\text{lin}}({x}_i).
\end{equation}
This preserves the energy information about the structure without a compositional degree of freedom, allowing energies of any level of theory to be compared. This process is also replicated for each model representation $f$ and its predicted energies, yielding the corresponding deviation from its own linear compositional model 
\begin{equation}
    \Delta E_f({x}_i) = \hat{E}_f({x}_i) - \hat{E}_{\text{lin}, f}({x}_i).
\end{equation}
where $\hat{E}_f$ denotes the model’s predicted energy and $\hat{E}_{\text{lin}, f}$ is the linear model trained on its predictions. The performance of each $\hat{E}_{\text{lin}}({x}_i)$ is shown in Fig. \ref{fig:all_model_linear_compositional_dof}

We then define the energy regression mean absolute error (MAE) for model representation $f$ as
\begin{equation}
    \text{MAE}_f = \frac{1}{N} \sum_{i=1}^N \left| \Delta E_{\text{true}}({x}_i) - \Delta E_f({x}_i) \right|,
\end{equation}
which quantifies the mean absolute deviation between each model’s corrected energy prediction and the ground truth deviation from compositionality. We plot $\Delta E_{\text{true}}({x}_i)$ against $\Delta E_f({x}_i)$ for all models in Fig. \ref{fig:deviations_from_linear_model_mae}. This formulation isolates the non-compositional, structural component of the energy landscape, enabling a direct comparison of models trained at different DFT levels of theory.


%

Acknowledgements are not compulsory. Where included they should be brief. Grant or contribution numbers may be acknowledged.

Please refer to Journal-level guidance for any specific requirements.

\backmatter

\section*{Declarations}

\subsection*{Acknowledgements}

We would like to thank Juno Nam, Antonia Kuhn, and Xiaochen Du for their early efforts, and Jinyeop Song, Lucas Pinede, and Matteo Carli for helpful discussions. The authors acknowledge the MIT SuperCloud and Lincoln Laboratory Supercomputing Center that have contributed to the research results reported within this paper.

\subsection*{Funding}
S.Y. thanks Ilju Foundation for their PhD fellowship support. S.E. recognizes MIT MGAIC for their funding support. 

\subsection*{Competing interests}
The authors declare no competing interests.

\subsection*{Data Availability}
All five datasets can be accessed through their respective citations.

\subsection*{Code Availability}
Code for extracting the embeddings from each of the 59 models and calculating each of the metrics will soon be available on the Learning Matter GitHub (\href{https://github.com/learningmatter-mit}{https://github.com/learningmatter-mit}).

\subsection*{Author Contribution}
S.Y. and R.G.B. led conceptualization of the project. S.Y. led planning of the methods, along with S.E. S.E. carried out computational experiments and visualized all analyses. S.E., S.Y., and R.G.B. prepared the original draft. All authors reviewed and approved the manuscript.

\bibliography{nature}
\bibliographystyle{sn-nature}

\newpage
\begin{appendices}

\section{Metrics}\label{si_metrics}

Here, we discuss the 4 different metrics we use in our analysis to quantify representational alignment through measuring latent space similarity between models. Each metric quantifies how similar a pair of models is. All metrics besides $I_d$ are bounded from 0 (random noise) to 1 (complete alignment). We define the following terms to help improve symbolic digestion, inspired by notation in \cite{huh2024prh}.
\begin{itemize}
    \item We define a \textbf{representation} as a function $f : \mathcal X \rightarrow \mathbb R^n$ that provides a vector of features for each inputted datapoint in some domain $\mathcal X$. 
    \item \textbf{Embeddings}, defined as $f(x_i)$ for selected $x_i \in \mathcal X$ ($1 \leq i \leq N$ samples), are thus the values of each model representation $f$. Embeddings are extracted as detailed in Section \ref{SI:models}. We can define a convenient matrix $\phi : \mathbb R^N \times \mathbb R^n$ as
    \begin{equation}
    \mathbf{\Phi}(\mathbf x) =
    \begin{bmatrix}
    –– & f(x_1)^\top & –– \\
    –– & f(x_2)^\top & –– \\
    &\vdots& \\
    –– & f(x_N)^\top & ––
    \end{bmatrix},
    \end{equation}
    where each row $i$ of the matrix is the embedding by representation $f$ of structure $x_i$. All embeddings $f(x_i)$ are normalized to have a maximum element value of 1, and similarly for each row of $\mathbf \Phi (\mathbf x)$
    \item A \textbf{kernel} $K : \mathcal X \times \mathcal X \rightarrow \mathbb R$ evaluates the distance between embeddings in representation space. Specifically, we can define the inner product kernel 
    \begin{equation}
        K_{f, g}(x_i, x_j) = \langle f(x_i) \cdot g(x_j) \rangle,
    \end{equation}
    which is the normalized dot product of the embeddings of structures $x_i, x_j \in \mathbb X$ by models $f$ and $g$. This yields convenient definitions for embedding self-similarity, $K_{f, f} = \mathbf{\Phi}(\mathbf x) \mathbf{\Phi}(\mathbf x)^\top $, and cross-modal similarity, $K_{f, g} = \mathbf{\Phi}(\mathbf x) \mathbf{\Psi}(\mathbf x)^\top $, using embedding matrices $\mathbf{\Phi}(\mathbf x)$ using representation $f$ and $\mathbf{\Psi}(\mathbf x)$ using representation $g$.
    \item \textbf{Kernel alignment metrics} $m : \mathcal K \times \mathcal K \rightarrow \mathbb R$ evaluate how similar two kernels are. Intuitively, these metrics evaluate how different two representations $f$ and $g$ are in terms of $K_{f, f}$, $K_{g, g}$, and/or $K_{g, f}$.
    \item \textbf{Model alignment} is how similar the latent spaces of two models are, and is measured by representational similarity by metrics in sections A.1 through A.5.
    \item \textbf{Global} alignment metrics consider a total of $2N$ embeddings  from representations $f$ and $g$ at a time. \textbf{Local} metrics consider far fewer embeddings, trading off increased sensitivity for decreased breadth. Intuitively, global metrics quantify how similar two entire representation spaces are to each other, while local metrics find the local neighborhoods around the same embedded data point in both latent spaces and compare them.
    \item Intuitively speaking, the representation space \textbf{manifold} can thought of as the shape of the latent space. All embeddings lie as points on this surface, and the complexity (how many twists and turns) the manifold has is directly related to how complex the representation is. 
\end{itemize}

\subsection{Centered kernel nearest-neighbor alignment (CKNNA)}\label{SI:CKNNA}

We use the Centered Kernel Nearest-Neighbor Alignment (CKNNA) metric for much of our primary analysis, as proposed by \cite{huh2024prh}. Intuitively, CKNNA is high if the local neighborhood of the same data point in two different representation spaces is identical, i.e., if both representations agree on which data points $(x_i, y_i)$ are the most similar. We now briefly reproduce its definition from the original work.

\subsubsection{Definition}

For models with different modalities, we define $(x_i, y_i) \in \mathcal X$ as samples from the multi-modal data distribution $\mathcal X$. As an example, $x_i$ could be an ASE Atoms object with 3D coordinates of each atom and $y_i$ could be the corresponding SMILES string. For models with the same modality, e.g. two machine learning interactive potentials (MLIPs), $x_i = y_i$. 

Next, using matrix $\mathbf{\Phi}(\mathbf x)$ from representation $f$ and $\mathbf{\Psi}(\mathbf y)$ from representation $g$, we define $K_{f, f}(x_i, x_j)$ ($K$ for short) and $L_{g, g}(y_i, y_j)$ ($L$ for short). The ``centered" versions of these inner product kernels are:
\begin{equation}
    \bar K_{ij} = \langle f(x_i), f(x_j) \rangle - \mathbb E \left[ \langle f(x_i), f(x_j) \rangle \right],
\end{equation}
and 
\begin{equation}
    \bar L_{ij} = \langle g(y_i), g(y_j) \rangle - \mathbb E \left[ \langle g(y_i), g(y_j) \rangle \right].
\end{equation}
The cross-covariance of $K$ and $L$ is given by 
\begin{equation}
    \frac{1}{(n - 1)^2}\text{Trace}(\bar K\bar L^\top) \approx \text{HSIC}(K, L),
\end{equation}
which is also an estimator of the Hilbert-Schmidt Independence Criterion \cite{hsic}. The Centered Kernel Alignment (CKA) measures the congruence of two random variables, is bounded to $[0, 1]$, is invariant to isotropic scaling, and offers a strict notion of alignment, thus globally measuring representational similarity between models \cite{cka}. It's defined as a normalized version of HSIC:
\begin{equation}
    \text{CKA}(K, L) = \frac{\text{HSIC}(K, L)}{\sqrt{\text{HSIC}(K, K)\ \text{HSIC}(L, L)}}.
\end{equation}
Lastly, we can restrict the definition of alignment to only consider cross-covariance measurements of the mutual nearest neighbors of a sample:
\begin{equation}
    \text{Align}(K, L) = \sum_{i, j} \alpha(i, j) \bar K_{ij} \bar L_{ij},
\end{equation}
where $\alpha(i, j)$ is 1 only if embedding $f(x_j)$ is in the $k$ nearest neighbors of $f(x_i)$ and if and $g(y_j)$ is in the $k$ nearest neighbors of $g(y_i)$, and is 0 otherwise.

Lastly, we define CKNNA as the normalized version of this metric:
\begin{equation}
    \text{CKNNA}(K, L) = \frac{\text{Align}(K, L)}{\sqrt{\text{Align}(K, K)\ \text{Align}(L, L)}}.
\end{equation}

\subsubsection{Justification and consistent globality via $k$}

\begin{figure}
    \centering
    \includegraphics[width=0.75\linewidth]{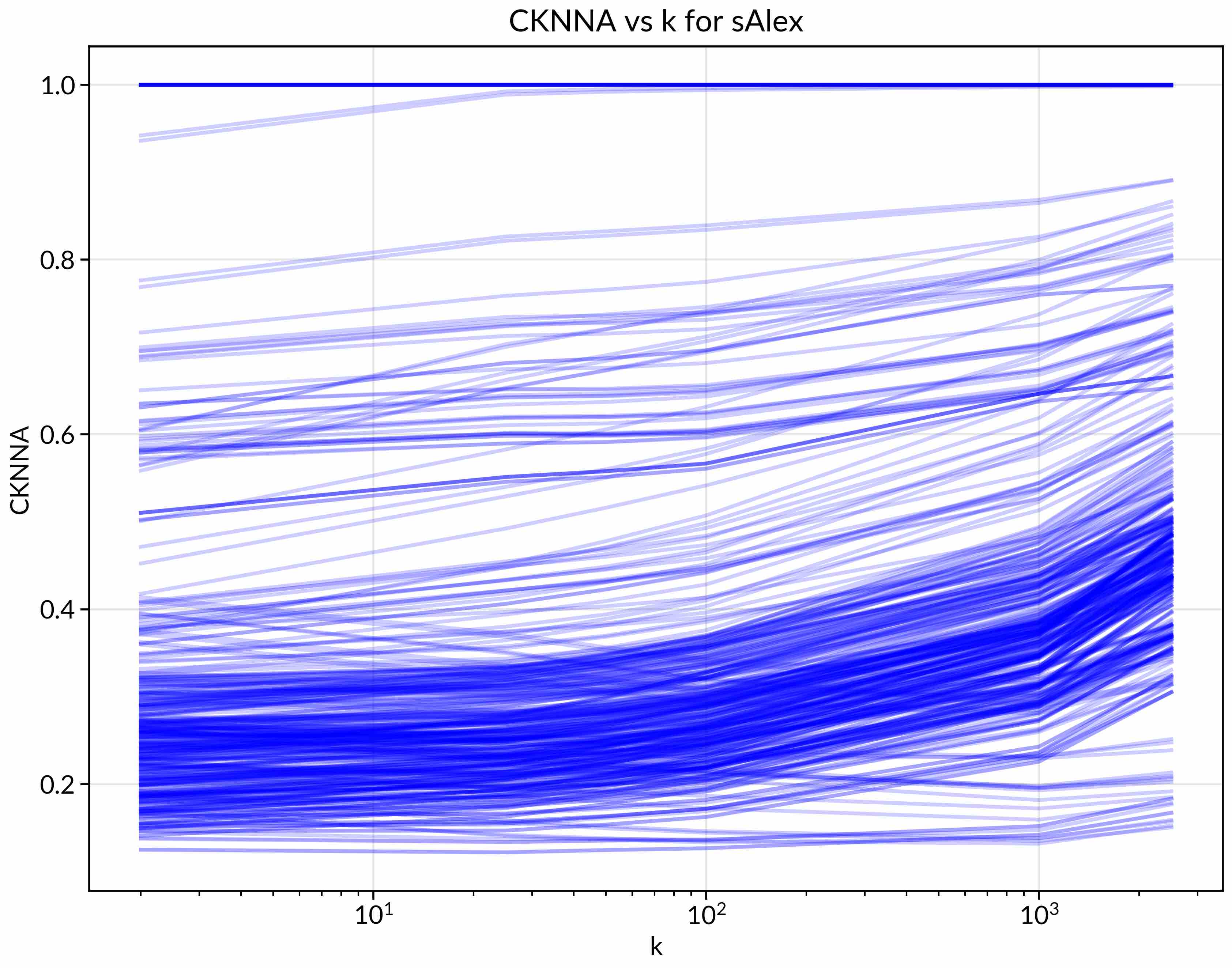}
    \caption{CKNNA's sensitivity to k evaluated with the sAlex dataset. \textbf{A} shows that as $k$ increases, CKNNA increases monotonically, preserving the ordering of model alignments almost exactly.}
    \label{fig:si:cknna_vs_k}
\end{figure}

We chose CKNNA because it can reveal alignment between models that is otherwise ignored by more global metrics (like those discussed in the following sections). Although our choice of $k = 25$ is far smaller than the total number of embeddings $N = 50,000$ for much of our analysis, we find that as $k$ increases, CKNNA yields consistent results, as shown in Fig. \ref{fig:si:cknna_vs_k}. By testing CKNNA with $k$ values of 2, 25, 50, 100, 1000, and 2500, we observe that CKNNA increases monotonically. More importantly, the trend of which models have higher or lower CKNNA is mostly consistent regardless of k. This shows that our results in the main text, although generated for $k = 25$, are also applicable on a global scale. 

\subsubsection{Constructing the evolutionary tree of models}\label{SI:phylogenetic}

We visualized the full CKNNA matrix (omitting the random model baselines) as a phylogenetic tree by using CKNNA similarities as proxies for evolutionary differences. Here, we describe the method used to visualize the tree. We first convert the pairwise CKNNA similarity matrix $\mathbf{C} \in \mathbb{R}^{M \times M}$, where $M$ is the number of models, into a distance matrix $\mathbf{D}$ suitable for phylogenetic analysis. Each entry $C_{ij}$ represents the degree of local alignment between models $f_i$ and $f_j$, such that $C_{ii} = 1$ and $0 \le C_{ij} \le 1$\cite{huh2024prh}.

\paragraph{Creating the symmetric distance matrix from CKNNAs.}
We first interpret $\mathbf{C}$ as a probabilistic confusion matrix and define two conditional confusion profiles:
\begin{align}
    R_i(j) &= P(\text{predicted model} = j \mid \text{true model} = i) 
    = \frac{C_{ij} + \varepsilon}{\sum_{j'} C_{ij'} + M\varepsilon}, \\
    K_i(j) &= P(\text{true model} = j \mid \text{predicted model} = i) 
    = \frac{C_{ji} + \varepsilon}{\sum_{j'} C_{j'i} + M\varepsilon},
\end{align}
where $\varepsilon$ is a small numerical constant ensuring numerical stability.

To obtain a symmetric measure of representational dissimilarity, we compute the Jensen--Shannon divergence (JSD) between these distributions:
\begin{equation}
    \mathrm{JSD}(p, q) = \tfrac{1}{2}\mathrm{KL}(p \,\|\, m) + \tfrac{1}{2}\mathrm{KL}(q \,\|\, m),
    \quad m = \tfrac{1}{2}(p + q),
\end{equation}
where $\mathrm{KL}(p \,\|\, q) = \sum_k p_k \log_2 \frac{p_k}{q_k}$ is the Kullback--Leibler divergence (in bits). The JSD is bounded to $[0,1]$, symmetric, and $\sqrt{\mathrm{JSD}}$ defines a metric in probability space.

We define the distance between models $i$ and $j$ as a convex combination of row- and column-profile divergences:
\begin{equation}
    D_{ij} = \alpha\, \sqrt{\mathrm{JSD}(R_i, R_j)} 
           + (1 - \alpha)\, \sqrt{\mathrm{JSD}(K_i, K_j)}, \quad D_{ii} = 0,
\end{equation}
where $\alpha \in [0,1]$ balances the relative emphasis between $R$ (who model $i$ is confused \textit{as}) and $K$ (who is confused \textit{as} model $i$). The resulting $\mathbf{D}$ thus encodes the effective representational dissimilarity between every pair of models.

\paragraph{Neighbor joining.}
We then construct an unrooted phylogenetic tree from $\mathbf{D}$ using the Neighbor Joining (NJ) algorithm \cite{saitou1987neighbor}. At each iteration, the pair $(i, j)$ minimizing the NJ criterion
\begin{equation}
    Q_{ij} = (M - 2) D_{ij} - \sum_k D_{ik} - \sum_k D_{jk}
\end{equation}
is selected for merging. Edge lengths are computed as
\begin{align}
    l_i &= \tfrac{1}{2}\left[D_{ij} + \frac{\sum_k (D_{ik} - D_{jk})}{M - 2}\right], \\
    l_j &= D_{ij} - l_i,
\end{align}
forming a new internal node $u$ with edges of length $l_i$ and $l_j$ to nodes $i$ and $j$, respectively. Distances from $u$ to the remaining nodes $k$ are updated according to
\begin{equation}
    D_{uk} = \tfrac{1}{2}(D_{ik} + D_{jk} - D_{ij}),
\end{equation}
and the procedure is repeated until only two nodes remain. The final topology is exported in Newick format for visualization.

The resulting tree provides an interpretable geometric visualization of cross-model alignment, where shorter branch lengths indicate stronger local agreement in CKNNA space and deeper bifurcations reflect divergence in learned representation manifolds across modalities and training paradigms.

\subsection{Intrinsic dimension ($I_d$)}

The intrinsic dimension $I_d$ of a representation quantifies how complex it is (the smallest number of dimensions needed to accurately represent a set of embeddings $\mathbf \Phi(\mathbf x)$. A simple way of estimating $I_d$ is to use principal component analysis (PCA) and find the smallest number of dimensions to still yield reasonable representations. Although this works well for embeddings that lie on linear data manifolds, estimating $I_d$ in more complicated regimes is an active research field. The TwoNN method is a fast and consistent method for doing so \cite{twonn}. Specifically, for each embedding $f(x_i)$, the distances to the two nearest neighbors $r_1(x_i)$ and $r_2(x_i)$ are calculated. The distribution $\mu = \frac{r_2(x_i)}{r_1(x_i)}$ roughly follows $f(\mu) = I_d \mu^{-I_d - 1}$ \cite{basile2025intrinsicdimensioncorrelationuncovering}. After generating the cumulative distribution function $P(\mu)$ by sorting the values of $\mu$ in ascending order and normalizing, we can calculate
\begin{equation}
    I_d(\mathbf{\Phi}(\mathbf x)) = -\frac{\ln(1 - P(\mu))}{\ln(\mu)}
\end{equation}
via linear regression.

There are also other methods for computing $I_d$ that consider more than just the two nearest neighbors, like the Maximum Likelihood Estimator (MLE) method \cite{mle}. We did not observe a significant difference in values of $I_d$ for $k = $ 5, 10, 25, 50, 100, 500, and 1000 using the MLE method, and we found that $I_d$ using MLE for our analysis agreed closely with $I_d$ using TwoNN. As a result, we believe our choice of $k = 50$ is appropriate trade off for globality and shorter analysis runtimes.

\subsection{Dimensionality correlation (dCor)}\label{si:dcor}
Distance correlation measures how strong the pairwise distances between points in one embedding space correspond to those in another, and considers all $2N$ total embeddings when comparing representation spaces $f$ and $g$. Intuitively, it generalizes Pearson correlation to high-dimensional, nonlinear relationships \cite{}. 

First, for embeddings $\mathbf{\Phi}(\mathbf{x})$, we can write $a_{k, l} = \|f(x_k) - f(x_l)\|_2$, $\bar a_{k, \cdot} =\frac{1}{N}\sum_{l=1}^N a_{k, l}$, $\bar a_{\cdot, l} =\frac{1}{N}\sum_{k=1}^N a_{k, l}$, and $\bar a_{\cdot, \cdot} = \frac{1}{N^2}\sum_{k, l = 1} ^N a_{k, l}$. Now, we define a centered distance metric $A_{k, l} = a_{k, l} - \bar a_{k, \cdot} - \bar a_{\cdot, l} + \bar a_{\cdot, \cdot}$. We can write similar quantities for $\mathbf{\Psi}(\mathbf{y})$ in terms of $g(y_i)$ such that $b_{k, l} = \|g(y_k) - g(y_l)\|_2$ and $B_{k, l} = b_{k, l} - \bar b_{k, \cdot} - \bar b_{\cdot, l} + \bar b_{\cdot, \cdot}$. Then, we can simply write
\begin{equation}
    \text{dCor}(\mathbf{\Phi}(\mathbf{x}), \mathbf{\Psi}(\mathbf{y})) = \frac{\text{cov}(A, B)}{\sqrt{\text{var}(A)\text{var}(B)}}.
\end{equation}
Intuitively, as dCor approaches 1, one embedding becomes a monotonic function of the other. 

\subsection{Information imbalance (II)}\label{si:ii}
Unlike all of the previous metrics, information imbalance (II) is an asymmetric measure that quantifies how much more information one representation has than another \cite{ii}. It's based on the idea that finding an embedding's nearest neighbors is much more informative than finding the points with the smallest L1 distance from the reference embedding in any given dimension. 

We first define $r^f_{ij}$, which is the nearest neighbor rank of $f(x_j)$ with respect to $f(x_i)$. As an example, if $f(x_j)$ was the second nearest neighbor of $f(x_i)$, $r^f_{ij}$ would be $2$. We then assemble a conditional rank distribution $p(r^g | r^f = 1)$, which is the probability distribution of the ranks $r^g_{ij}$ where each pair of embeddings $f(x_i)$ and $f(x_j)$ are nearest neighbors in $f$. The closer $p(r^g | r^f = 1)$ is to a delta function peaked at 1, the more information about representation $g$ is completely contained within representation $f$. The authors make this rigorous by defining a copula variable, or cumulative distribution, $c_f = \int_0 ^{d_f} p_f(w|x)dw$, where $p_f(w|x)$ is the probability of sampling a data point within distance $w$ from $x$. This can be conveniently estimated by $c_f \approx \frac{r^f}{N}$. Therefore, we can define the information imbalance between representation $f$ and $g$ as:
\begin{equation}
    \Delta(f \rightarrow g) = 2 \lim_{\epsilon \rightarrow0} \langle c_g | c_a = \epsilon \rangle.
\end{equation}

By measuring and plotting $\Delta(f \rightarrow g)$ and $\Delta(g \rightarrow f)$, which are not the same because $r^f_{ij} \neq r^g_{ij}$, we can learn about which models are more informative than others. Specifically, if $\Delta(f \rightarrow g) = 0$ and $\Delta(f \rightarrow g) = 0$ (towards the bottom left of an II plot), then the representations have learned completely identical information. If $\Delta(f \rightarrow g) = 1$ and $\Delta(f \rightarrow g) = 1$ (towards the top right of an II plot), then they have learned completely separate and orthogonal information. However, if $\Delta(f \rightarrow g) = 0$ and $\Delta(f \rightarrow g) = 1$, then representation $g$ is completely contained within representation $f$ (towards the top left/away from the diagonal in an II plot). If $\Delta(f \rightarrow g) \approx \Delta(f \rightarrow g) \sim 0.5$, then the representations share some information. 

The rank distributions $r^f_{ij}$ are sensitive to $k$ chosen for nearest neighbor search. As $k$ increases, nearest neighbor similarity decreases, and II yields that representation spaces learn increasingly orthogonal information, as shown in the information imbalance plots in the next section.

\section{Models}\label{SI:models}

We provide additional figures for evaluating model energy regression performance and LLM details for model configurations below.

\subsection{Energy regression MAE}

Here, we show two figures that elucidate our method for calculating energy regression MAE. Fig. \ref{fig:all_model_linear_compositional_dof} shows the performance of each linear compositional model fit to each model's energy predictions. 
We show the deviations of each model's energy predictions from their respective linear models of composition in Fig. \ref{fig:deviations_from_linear_model_mae}.
Lastly, we show the calculated energy regression MAEs with our method for each model on OMat24 structures in Fig. \ref{fig:omat24_maes}.

\afterpage{%
\clearpage
\begingroup
\begin{figure}[p]
    \centering
    \makebox[\textwidth]{\includegraphics[width=1.3\linewidth]{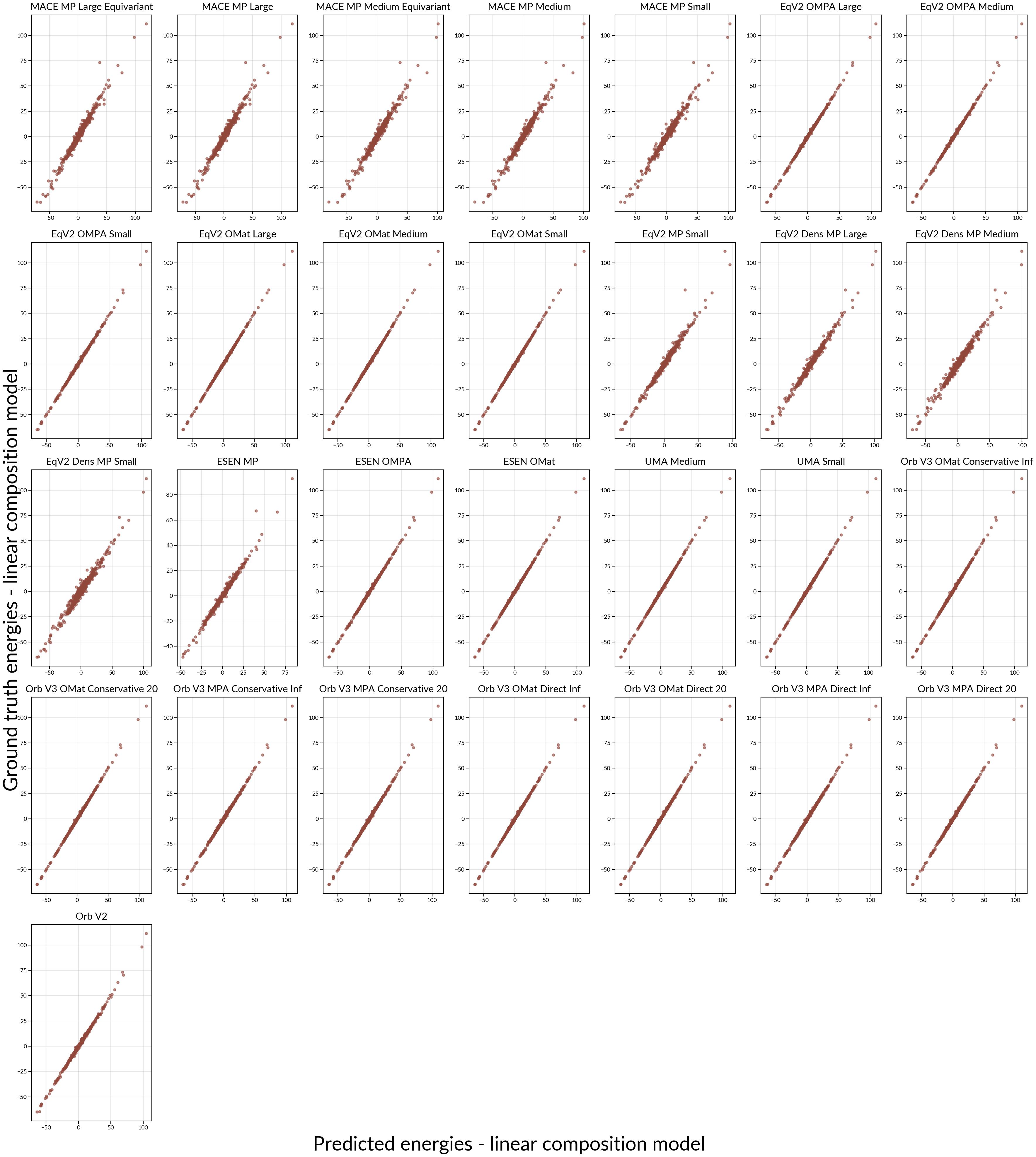}}
    \caption{Predictions of each model's energy outputs by a linear function of material composition versus the model's actual energy predictions.}
    \label{fig:all_model_linear_compositional_dof}
\end{figure}
\clearpage
\endgroup
}

\afterpage{%
\clearpage
\begingroup
\addtolength{\footskip}{3cm} 
\begin{figure}[htp!]
    \centering
    \makebox[\textwidth]{\includegraphics[width=1.3\linewidth]{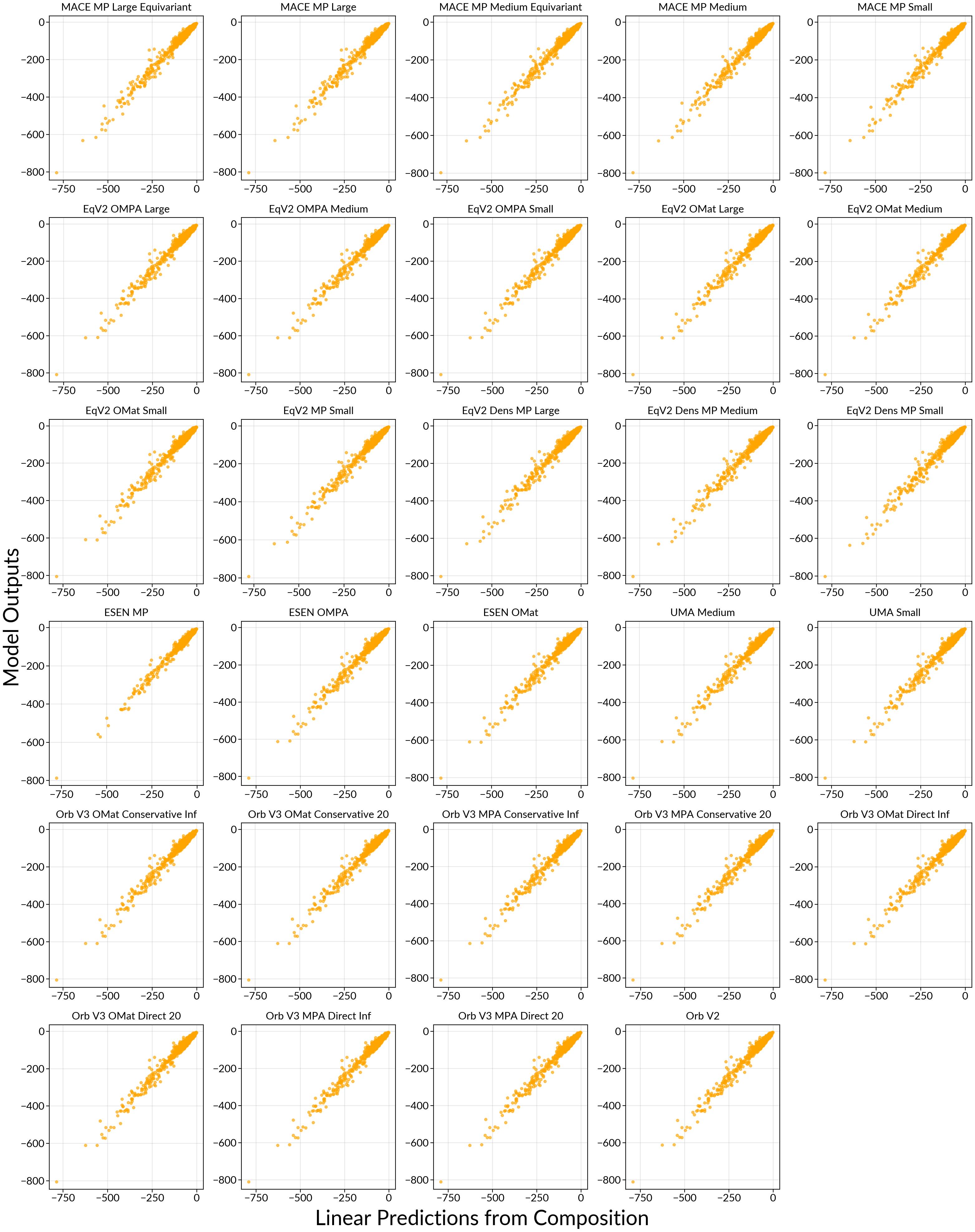}}
    \caption{The deviations of each model's energy predictions from their linear compositional model outputs.}
    \label{fig:deviations_from_linear_model_mae}
\end{figure}
\clearpage
\endgroup
}

\afterpage{%
\clearpage
\begingroup
\begin{figure}
    \centering
    \makebox[\textwidth]{\includegraphics[width=1.3\linewidth]{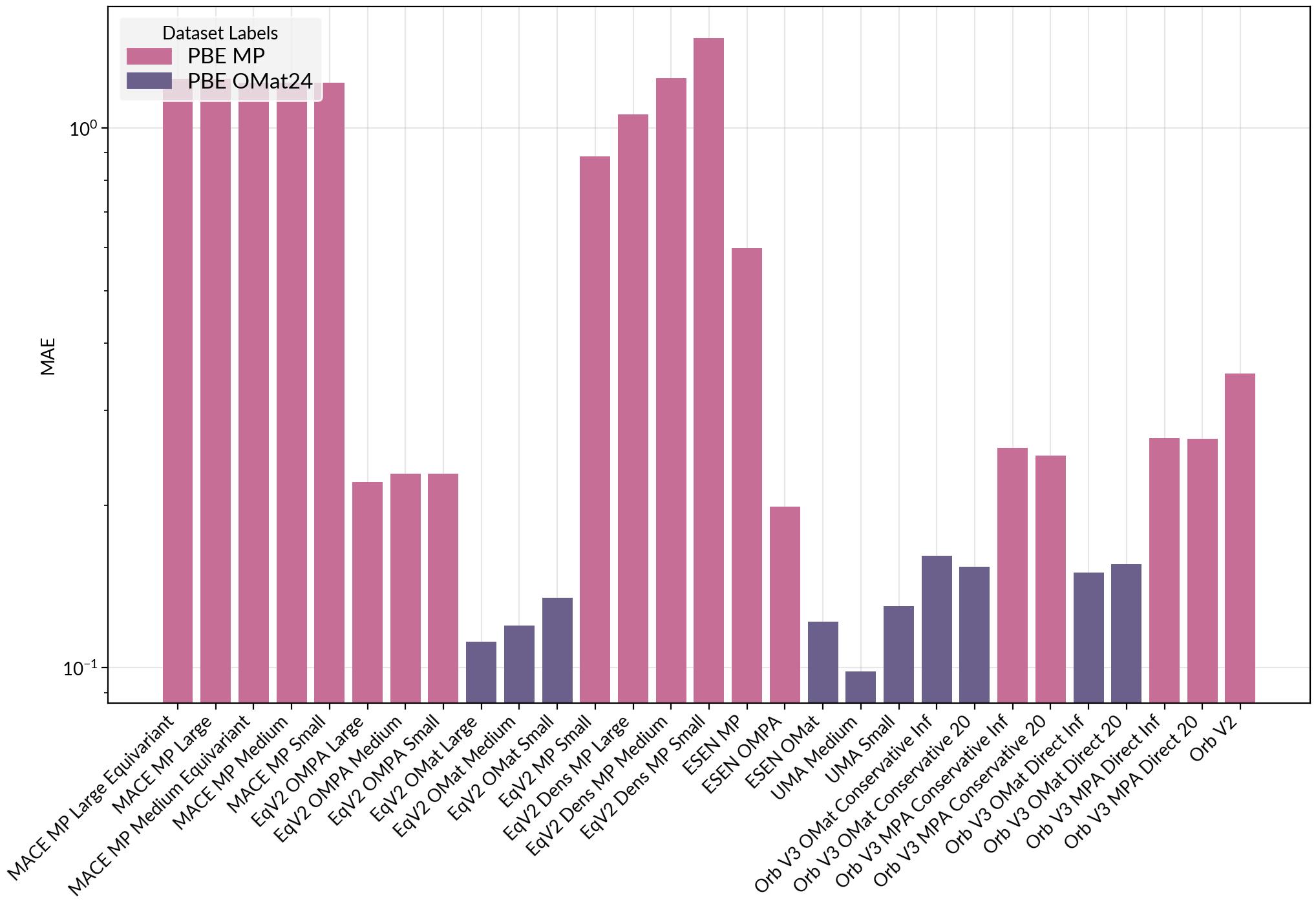}}
    \caption{Energy regression MAEs for each model, colored by the DFT level of theory they were trained on.}
    \label{fig:omat24_maes}
\end{figure}
\clearpage
\endgroup
}

\subsection{LLM system prompts}
Below is the default LLM system prompt.
\newpage
\begin{tcolorbox}[title=Default System Prompt]
You are a domain-agnostic scientific encoder. \\
    Your role is to interpret inputs such as:\\
    - Small molecules described as SMILES or SELFIES strings,\\
    - Proteins described as amino acid sequences,\\
    - Materials described as normalized structural text (formula, space group, lattice parameters, and fractional coordinates).\\

    You must focus only on the chemical, structural, and biological meaning of the inputs. \\
    Do not generate explanations, commentary, or unrelated text. \\
    Instead, represent the inputs in a consistent and semantically meaningful latent space, where structurally similar molecules, proteins, and materials map to nearby embeddings.\\

    Your interpretation should:\\
    - Treat SMILES/SELFIES as molecular graphs,\\
    - Treat protein sequences as biopolymers with ordered residues,\\
    - Treat material schemas as periodic crystals with lattice + sites,\\
    - Ignore formatting differences (whitespace, capitalization, punctuation) that don’t affect meaning,\\
    - Emphasize underlying chemistry, composition, and structure.\\

    Output should remain in hidden state embeddings only, not text. 
\end{tcolorbox}

Below is the ``Blank" LLM system prompt.
\begin{tcolorbox}[title=Blank LLM system prompt]
You are a domain-agnostic scientific encoder.
\end{tcolorbox}

\subsection{LLM example inputs}

Below is an example default LLM input:
\begin{tcolorbox}[title=Default LLM input]
SMILES: C[C@]12OC[C@H]1[C@H]1OC[C@H]12
\end{tcolorbox}

Below is an example ``Atomistic" model input:
\begin{tcolorbox}[title=Atomistic model input]
Molecule ASE atoms positions and information: Atoms(symbols='CH3C2H3OHC2HOCOH4', pbc=False)
{'numbers': [6, 1, 1, 1, 6, 6, 1, 1, 1, 8, 1, 6, 6, 1, 8, 6, 8, 1, 1, 1, 1], 'positions': [[-2.0406705094, -1.1742161595, -0.5740138071], [-1.4459254896, -2.0787833996, -0.4687043992], [-2.2475399162, -1.0132034606, -1.6279709632], [-2.9807740345, -1.2999713868, -0.0424259676], [-1.2779879231, 0.0198534811, 0.0098548837], [-2.0289281319, 1.3224133045, -0.2573246053], [-2.0775054027, 1.5191879876, -1.3240793916], [-3.0369983817, 1.2387729694, 0.1386739811], [-1.5339482034, 2.1513908963, 0.2393747589], [-1.1512597147, -0.1137750646, 1.4107771122], [-0.6669990939, -0.9325832833, 1.5762516748], [0.1329400021, 0.0672643768, -0.6224615909], [1.0280308312, 1.1964694506, -0.0878955818], [1.0537718111, 2.0798665429, -0.7215445844], [2.3135946221, 0.6139369749, -0.0816293663], [2.0948204176, -0.7102416847, 0.2774267155], [0.8625018358, -1.1100919018, -0.2767385903], [2.9009569741, -1.3323284588, -0.1239200534], [2.0529727397, -0.8205420561, 1.3797900024], [0.7191296381, 1.4600964251, 0.9334911515], [0.0482354107, 0.1116012524, -1.7162024649]], 'cell': [[0.0, 0.0, 0.0], [0.0, 0.0, 0.0], [0.0, 0.0, 0.0]], 'pbc': [False, False, False]}
\end{tcolorbox}

\subsection{LLM example generated text}

Embeddings of inputted text are extracted from the last hidden states output, before autoregressive generation. These are examples of the text generated by each LLM when provided SMILES input if they were to generate text each time.

\begin{tcolorbox}[title=Qwen3 30B model output with blank system prompted given smiles string (with thinking)]
I need to interpret this as a chemical structure.\\

First, I'll recall that SMILES (Simplified Molecular Input Line Entry System) is a notation for describing the structure of chemical molecules using short strings of characters.\\

Let me break down the SMILES string:\\

- "C" - a carbon atom\\
- "[C@]12" - a carbon atom with a stereochemistry specification (@) and ring closure numbers 1 and 2\\
- "O" - an oxygen atom\\
- "[C@H]1" - a carbon atom with stereochemistry (@) and hydrogen (H), with ring closure number 1\\
- "[C@H]1" - another carbon atom with stereochemistry (@) and hydrogen (H), with ring closure number 1\\
- "OC" - an oxygen atom followed by a carbon atom\\
- "[C@H]12" - a carbon atom with stereochemistry (@) and hydrogen (H), with ring closure numbers 1 and 2\\

Let me try to parse this more carefully. The ring closure numbers (1, 2) indicate that atoms are connected to form rings.\\

Let me try to draw this structure mentally:\\

1. Start with "C" - a methyl group (CH3-)\\
2. Then "[C@]12" - a chiral carbon that is part of two rings (ring 1 and ring 2)\\
3. Then "O" - an oxygen atom\\
4. Then "[C@H]1" - a chiral carbon with a hydrogen, part of ring 1\\
5. Then "[C@H]1" - another chiral carbon with a hydrogen, part of ring 1\\
6. Then "OC" - an oxygen atom followed by a carbon atom\\
7. Then "[C@H]12" - a chiral carbon with a hydrogen, part of rings 1 and 2\\

This seems a bit confusing. Let me try to interpret it more systematically.\\

In SMILES, ring closure numbers are used to indicate where a ring closes. So, when a number appears in brackets like [C@]1, it means that this atom is connected to the atom with the same number later in the string.\\

Let me try to parse it step...
\end{tcolorbox}

Here's the same model's output when provided with 3D atom positions and coordinates, leading to near-noise representations. 
\begin{tcolorbox}[title=Qwen3 30B model output with blank system prompted given 3D atom coordinates (with thinking)]
The user has provided a SMILES string and a corresponding atomic structure in a specific format. Let me analyze this.\\

First, let's look at the SMILES string: "C3HCH2OCHCHCH2OH3"\\

This appears to be a SMILES notation for a chemical compound, but it's not standard. Let me try to interpret it:\\

- "C3" typically means a chain of 3 carbon atoms\\
- "HCH2" would be a hydrogen attached to a carbon, then a CH2 group\\
- "O" is oxygen\\
- "CHCHCH2" would be a chain of three carbons with hydrogens\\
- "OH3" would be an OH group with three hydrogens, which doesn't make sense (should be OH)\\

This SMILES string seems to be malformed. A standard SMILES string would have proper connectivity and atom counts. For example, a common molecule with similar notation might be something like "COC" for methoxyethane, but this doesn't match.\\

Now, let's look at the atomic structure data:\\

- 'numbers': [6, 6, 6, 1, 6, 1, 1, 8, 6, 1, 6, 1, 6, 1, 1, 8, 1, 1, 1]\\
  - These are atomic numbers: 6 = Carbon, 1 = Hydrogen, 8 = Oxygen\\
  - Counting: 9 Carbons (6s), 9 Hydrogens (1s), 2 Oxygens (8s)\\
  - Total atoms: 9 + 9 + 2 = 20 atoms\\

- 'positions': A list of 19 3D coordinates (x, y, z)\\
  - Wait, there are 19 positions listed, but the 'numbers' list has 19 elements. So it's a 19-atom structure.\\

- 'cell': [[0.0, 0.0, 0.0], [0.0, 0.0, 0.0], [0.0, 0.0, 0.0]]\\
  - This indicates no periodic boundary conditions (pbc=False as specified)\\

- 'pbc': [False, False, False]\\
  - Confirms no periodic boundary conditions\\

Let me count the atoms in the 'numbers' list:\\
- 6 (C): 9 times\\
- 1 (H): 9 times\\
- 8 (O): 2 times\\

So it's a molecule with 9 carbons, 9 hydrogens, and 2 oxygens...



\end{tcolorbox}

\section{Datasets}\label{si_datasets}

Here, we overview each datasets and provide full CKNNA representational alignment matrices and information imbalance plots where referenced in the main text. All CKNNA, $I_d$, dCor, or information imbalance results for any of the five datasets are available upon request, and examples of each are shown for the representations of structures from the QM9 dataset.

\begin{figure}
    \centering
    \includegraphics[width=\linewidth]{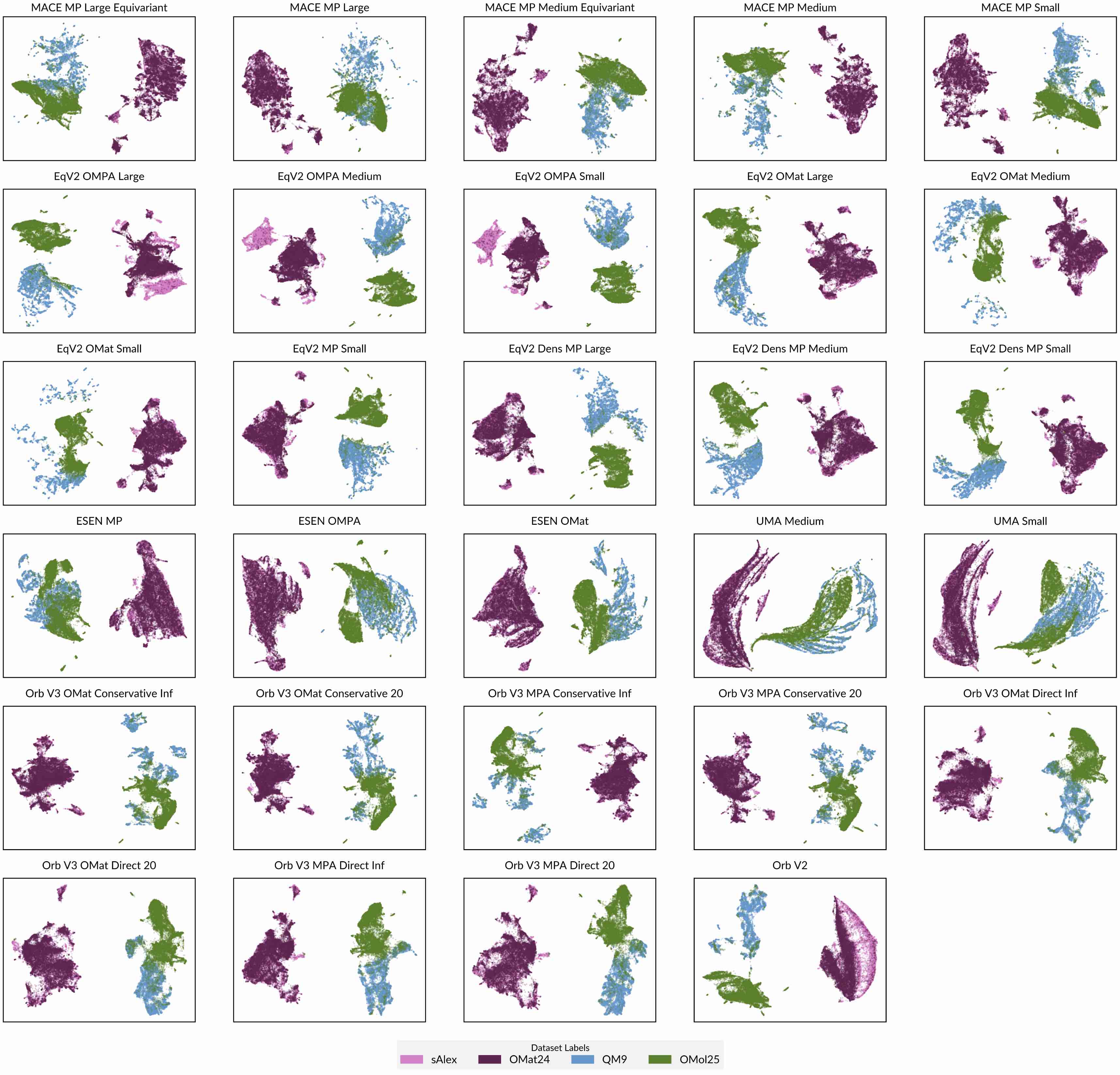}
    \caption{Model representations of 15,000 structures from each sAlex, OMat24, QM9, and OMol25, visualized in two dimensions with UMAP.}
    \label{fig:umap_all}
\end{figure}

\subsection{QM9}
The QM9 \cite{butler2018qm9} dataset contains roughly 134,000 small organic molecules containing up to nine heavy atoms (C, H, N, O, F). Each molecule in their gas-phase in vacuum was calculated with the B3LYP/6-31G(2df,p) DFT level of theory, labeled with total potential energies, forces, orbital energies (HOMO/LUMO), dipole moments, vibrational frequencies, polarizabilities, etc. The dataset provides SMILES strings (which were canonicalized before being passed into SMILES-based models), SELFIES strings (which can be generated from SMILES), and 3D atomic coordinates. 

\begin{figure}
    \centering
    \includegraphics[width=\linewidth]{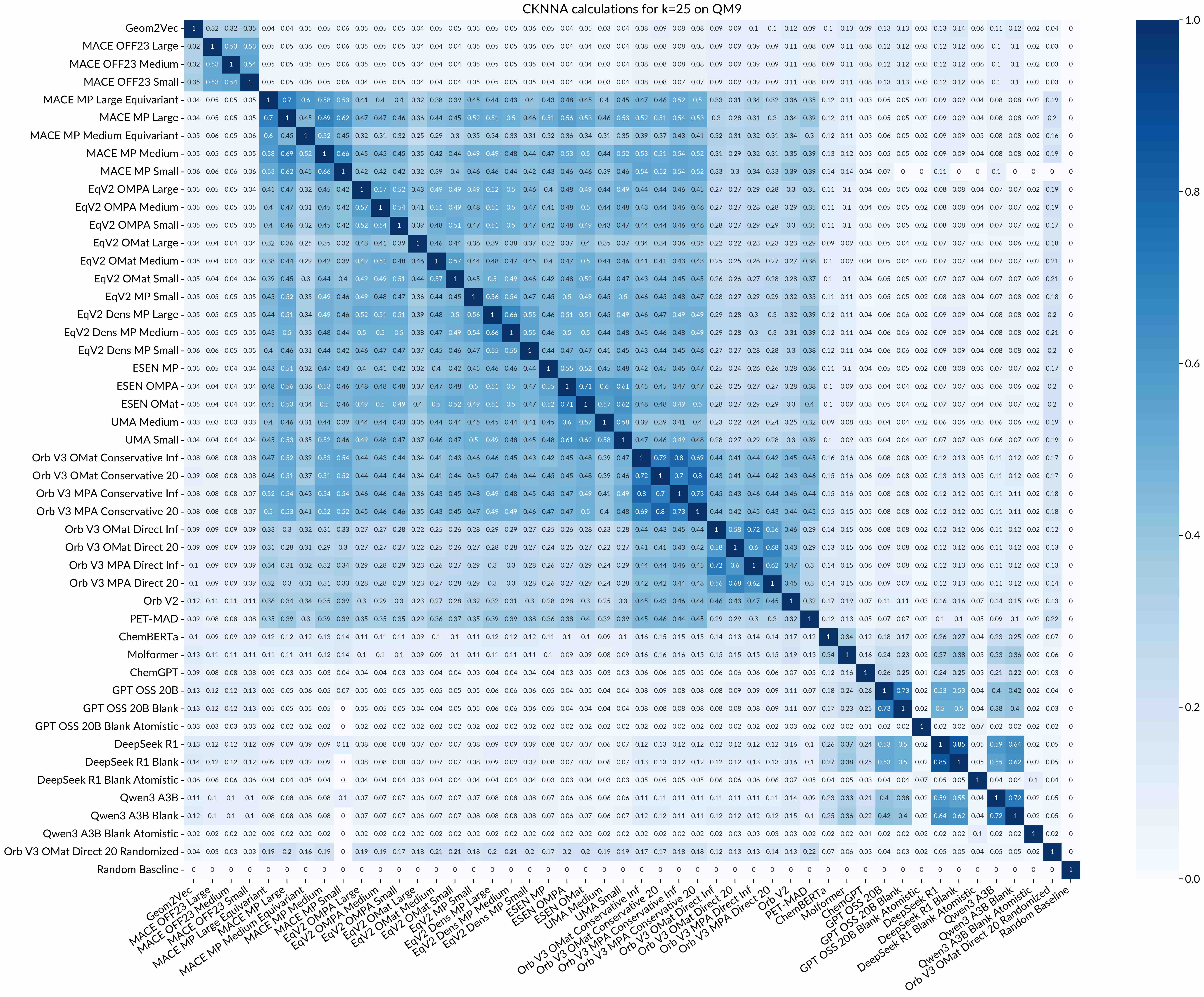}
    \caption{Full CKNNA correlation matrix ($k=25$) between each model's embeddings of the same 50,000 structures randomly sampled from QM9. A random baseline was also included.}
    \label{fig:qm9_full_cknna}
\end{figure}
\begin{figure}
    \centering
    \includegraphics[width=\linewidth]{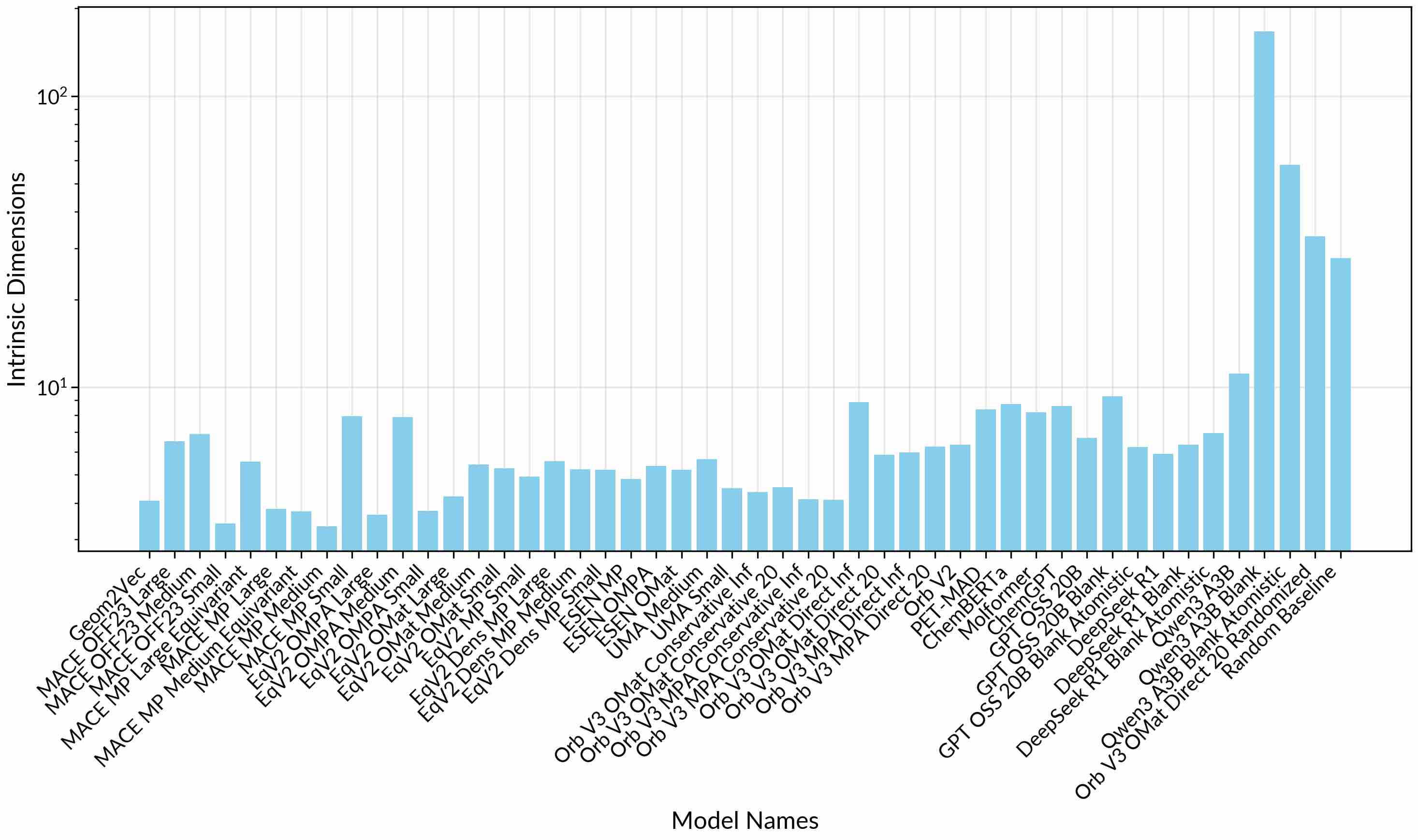}
    \caption{Intrinsic dimensions calculated for each model's embeddings ($k=50$) of structures from QM9. Of note, the intrinsic dimensions of random embeddings and LLMs are much higher than those of models trained on materials and small molecules.}
    \label{fig:qm9_id}
\end{figure}
\begin{figure}
    \centering
    \includegraphics[width=\linewidth]{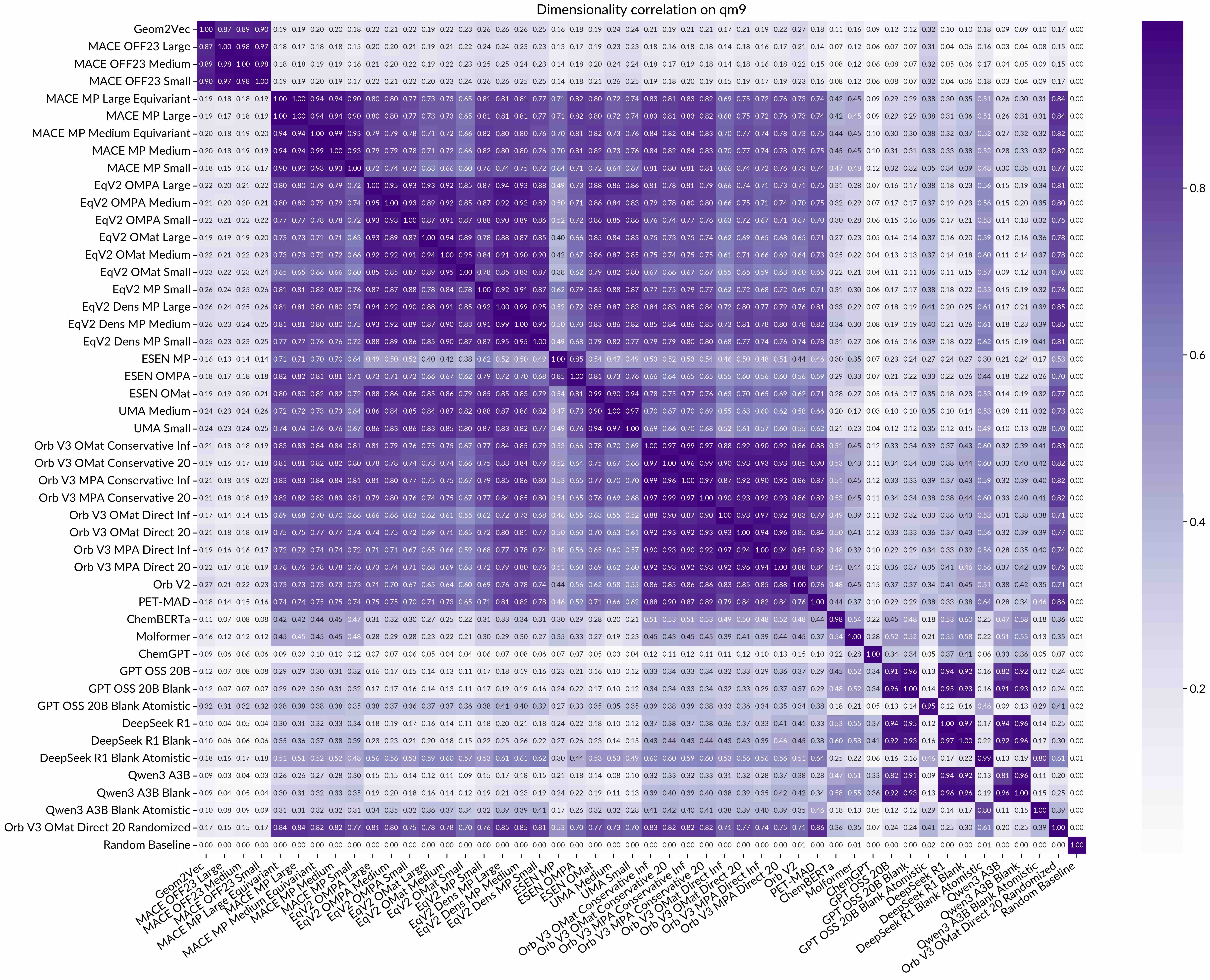}
    \caption{Distance correlation between each model's embeddings of structures from QM9.}
    \label{fig:qm9_dcors}
\end{figure}
\begin{figure}
    \centering
    \includegraphics[width=\linewidth]{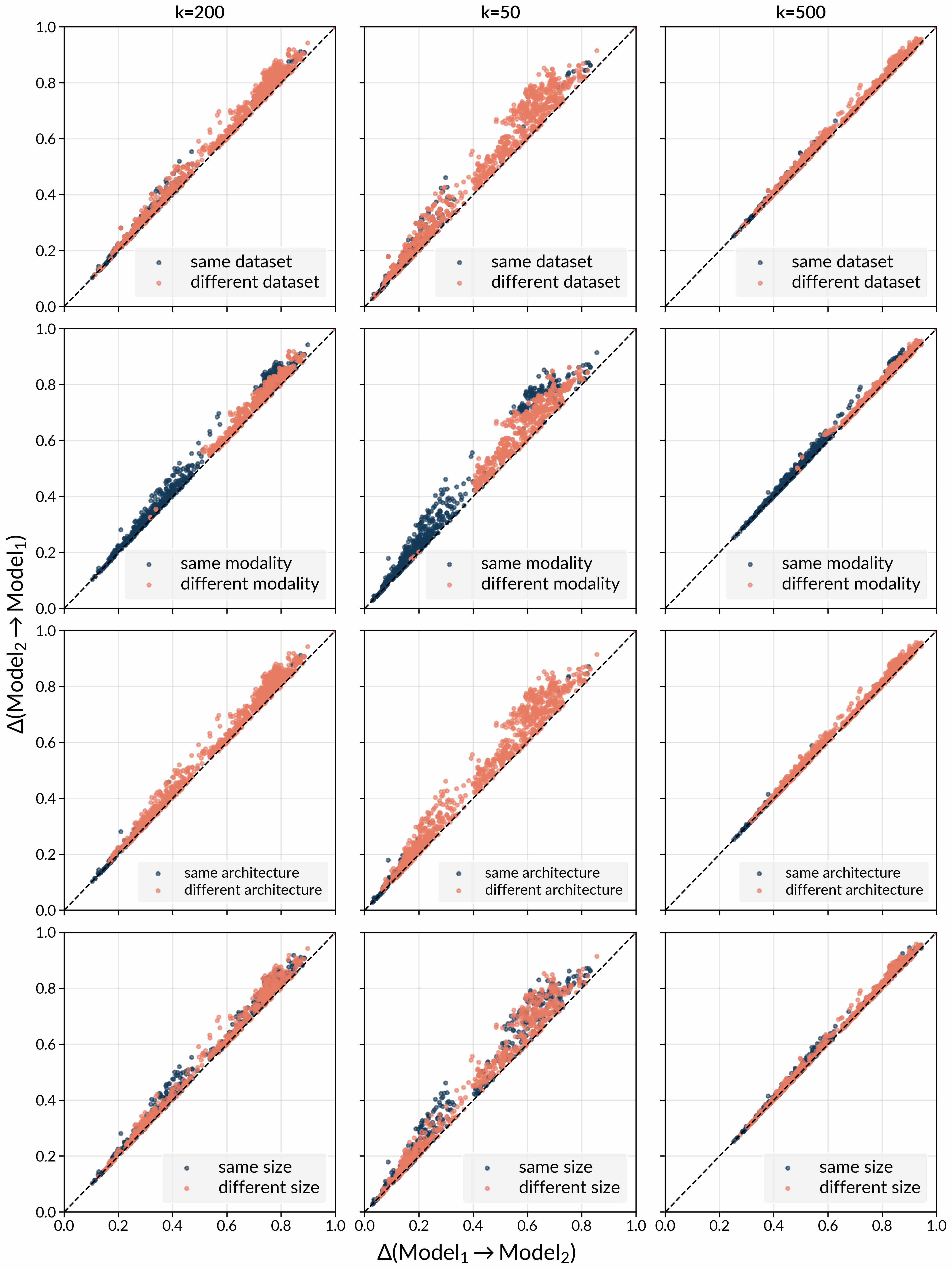}
    \caption{Information imbalance plots for embeddings of structures from QM9 at $k=50$, $k=200$, and $k=500$. Points are colored by whether or not the two models being compared have the same aspect (e.g. same dataset) or not. This provides additional intuition for which pairs of models learn similar information.}
    \label{fig:qm9_ii_by_aspect}
\end{figure}

\subsection{OMat24}
The (OMat24) \citep{barroso_omat24} dataset is a large-scale compilation of over 118 million inorganic materials structures. It includes both equilibrium and far-from-equilibrium configurations generated via high-throughput DFT workflows using PBE and PBE+U exchange–correlation functionals. OMat24 extends the Alexandria dataset by systematically applying atomic perturbations and structural transformations to yield dynamically diverse and non-equilibrium geometries. Each sample includes 3D atomic positions, total potential energies, forces, stress tensors, and compositional information. OMat24 underpins much of the recent progress in large materials foundation models, supporting learning across stability, energy, and force landscapes.

\begin{figure}
    \centering
    \includegraphics[width=\linewidth]{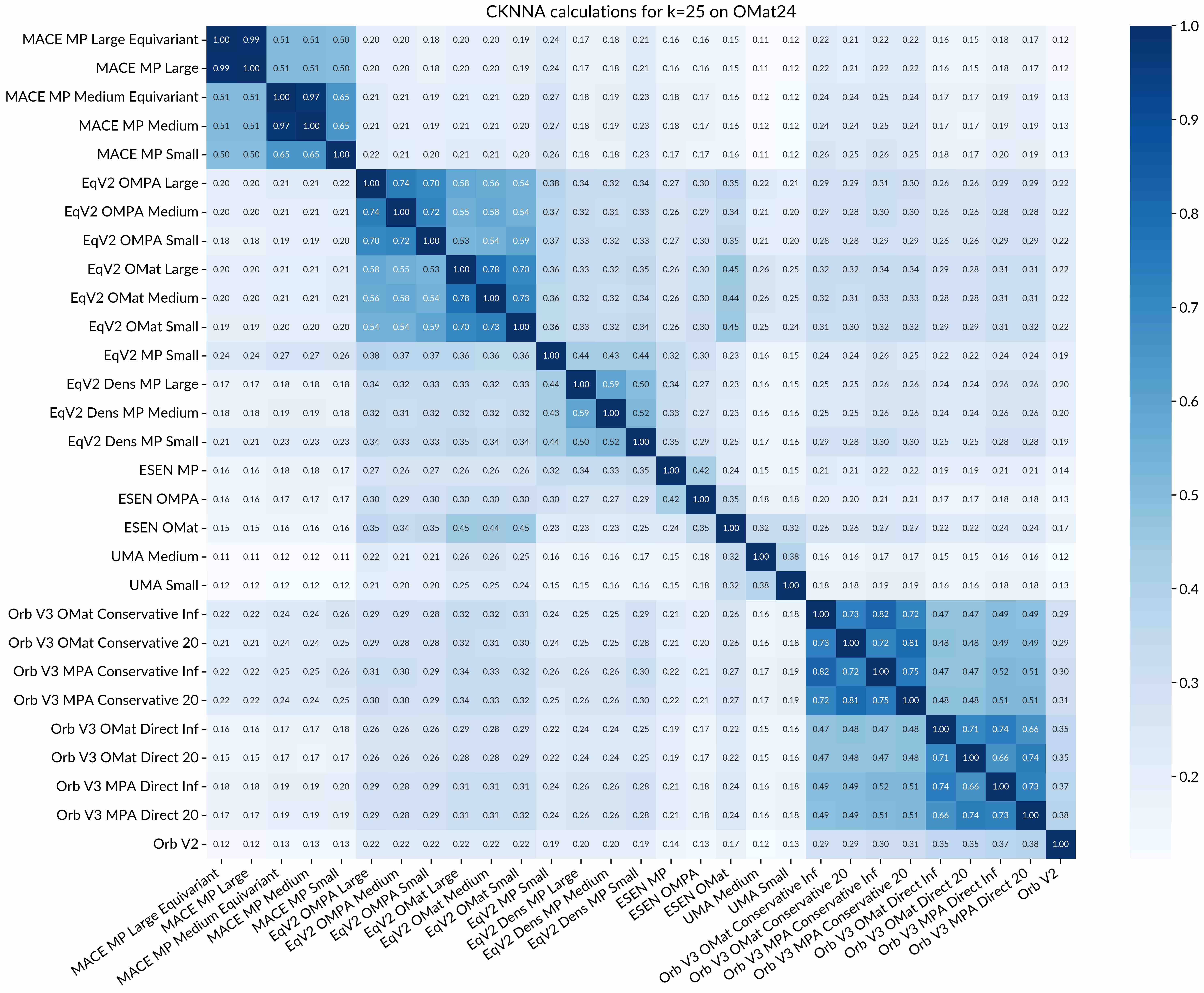}
    \caption{Full CKNNA correlation matrix ($k=25$) between each model's embeddings of the same 50,000 structures randomly sampled from OMat24.}
    \label{fig:OMat24_cknna}
\end{figure}

\subsection{sAlex}
The sAlex dataset is a subsampled, diversity-optimized subset of the Alexandria dataset, which comprises approximately ten million inorganic material structures at equilibrium computed using DFT. The original Alexandria database spans bulk, two-dimensional, and low-dimensional inorganic compounds, primarily at PBE or PBE+U levels of theory. Each entry includes 3D atomic coordinates, lattice parameters, and composition vectors describing the crystal stoichiometry. In this work, sAlex is considered an in-distribution dataset for materials foundation models, serving as a representative benchmark for crystalline solids with reduced redundancy and broad compositional coverage. 
\begin{figure}
    \centering
    \includegraphics[width=\linewidth]{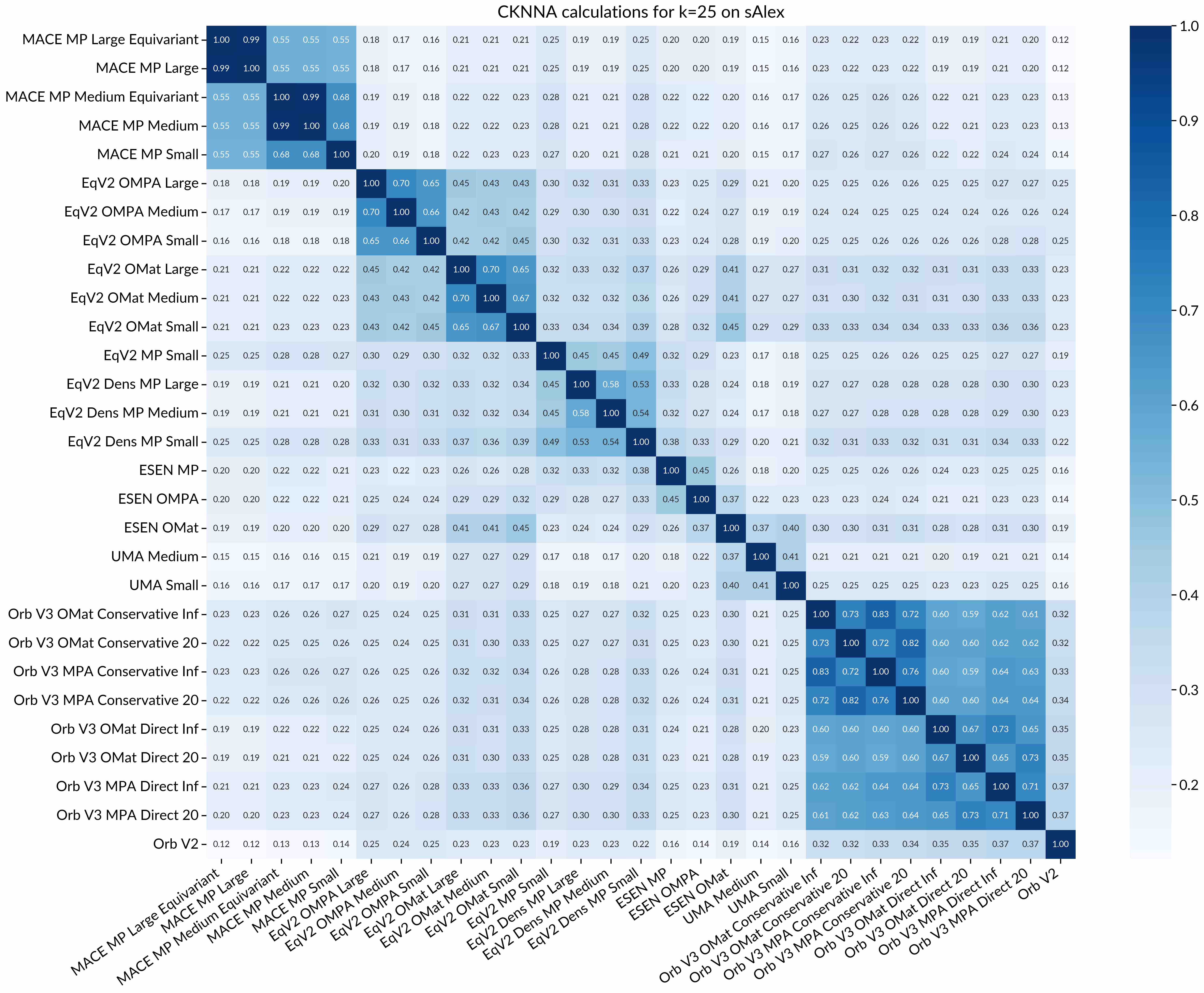}
    \caption{Full CKNNA correlation matrix ($k=25$) between each model's embeddings of the same 50,000 structures randomly sampled from sAlex.}
    \label{fig:sAlex_cknna}
\end{figure}

\subsection{OMol25}
The Open Molecules~2025 (OMol25)~\citep{levine2025openmolecules2025omol25} dataset contains over one hundred million molecular structures spanning organic, biomolecular, electrolyte, and transition-metal systems. Each entry is computed at the $\omega$B97M-V/def2-TZVPD level of DFT, a hybrid meta-GGA functional with range separation and empirical dispersion corrections, yielding high-accuracy potential energy surfaces and molecular properties. The dataset includes 3D atomistic positions and quantum-mechanical observables such as total energy, dipole moments, and charges. Due to its scale and inclusion of large and complex molecules, OMol25 serves as an out-of-distribution dataset relative to most existing molecular foundation models.

\begin{figure}
    \centering
    \includegraphics[width=\linewidth]{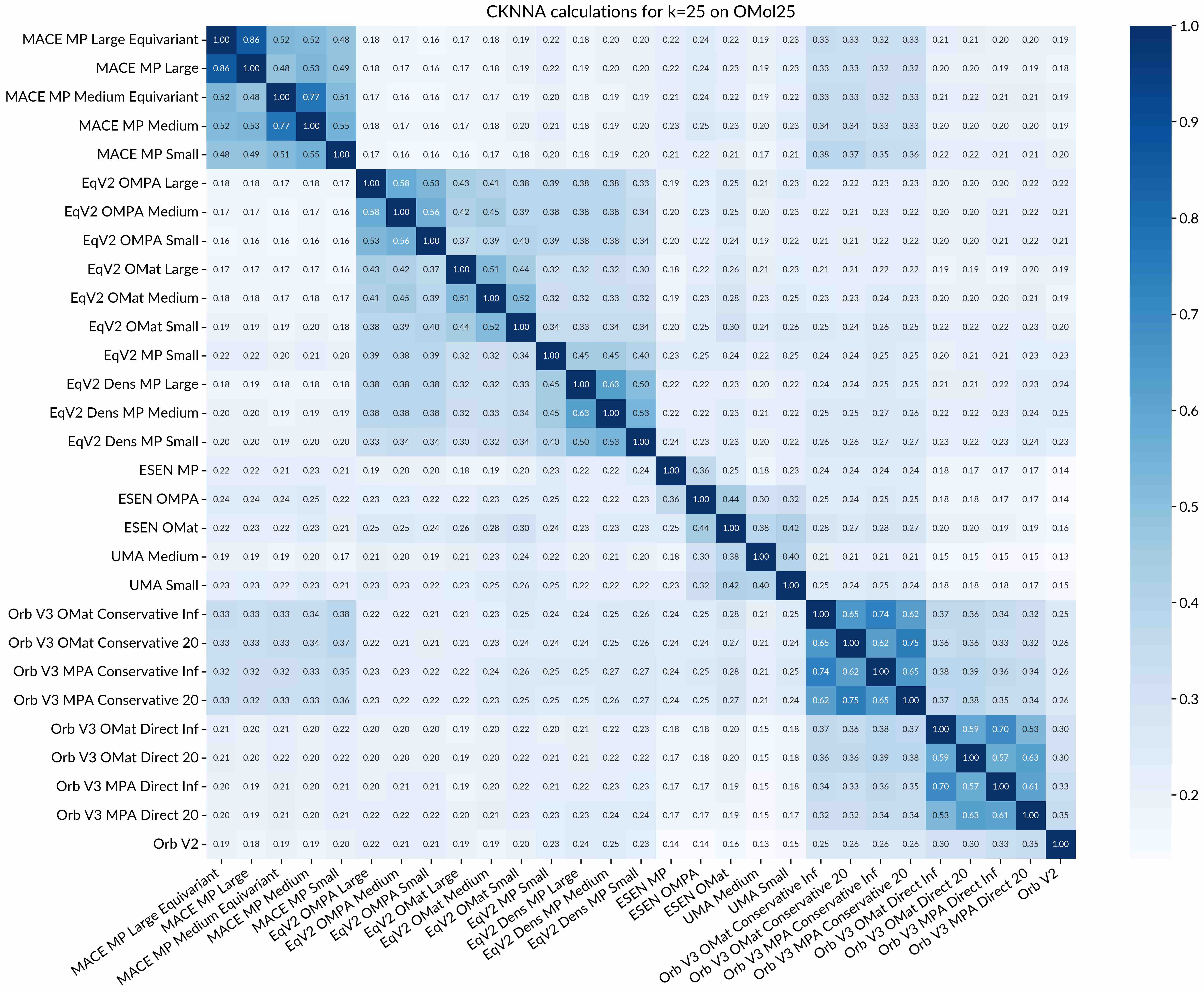}
    \caption{Full CKNNA correlation matrix ($k=25$) between each model's embeddings of the same 50,000 structures randomly sampled from OMol25.}
    \label{fig:OMol25_cknna}
\end{figure}

\subsection{RCSB}
The RCSB Protein Data Bank (PDB) \cite{rcsb} is the largest open-access repository of experimentally determined three-dimensional structures of biological macromolecules, containing over two hundred thousand protein, nucleic acid, and complex structures. Each entry is derived from experimental techniques like X-ray crystallography, nuclear magnetic resonance (NMR) spectroscopy, and cryo-electron microscopy (cryo-EM), providing atomic-resolution structural data with experimentally validated coordinates. The dataset includes 3D atomic positions, amino acid sequences, secondary structure annotations, ligand binding information, and crystallographic metadata such as resolution and R-factors. Due to its diversity of protein families, functional classes, and structural motifs, the RCSB PDB serves as a comprehensive benchmark for evaluating protein structure and sequence foundation models.

\begin{figure}
    \centering
    \includegraphics[width=\linewidth]{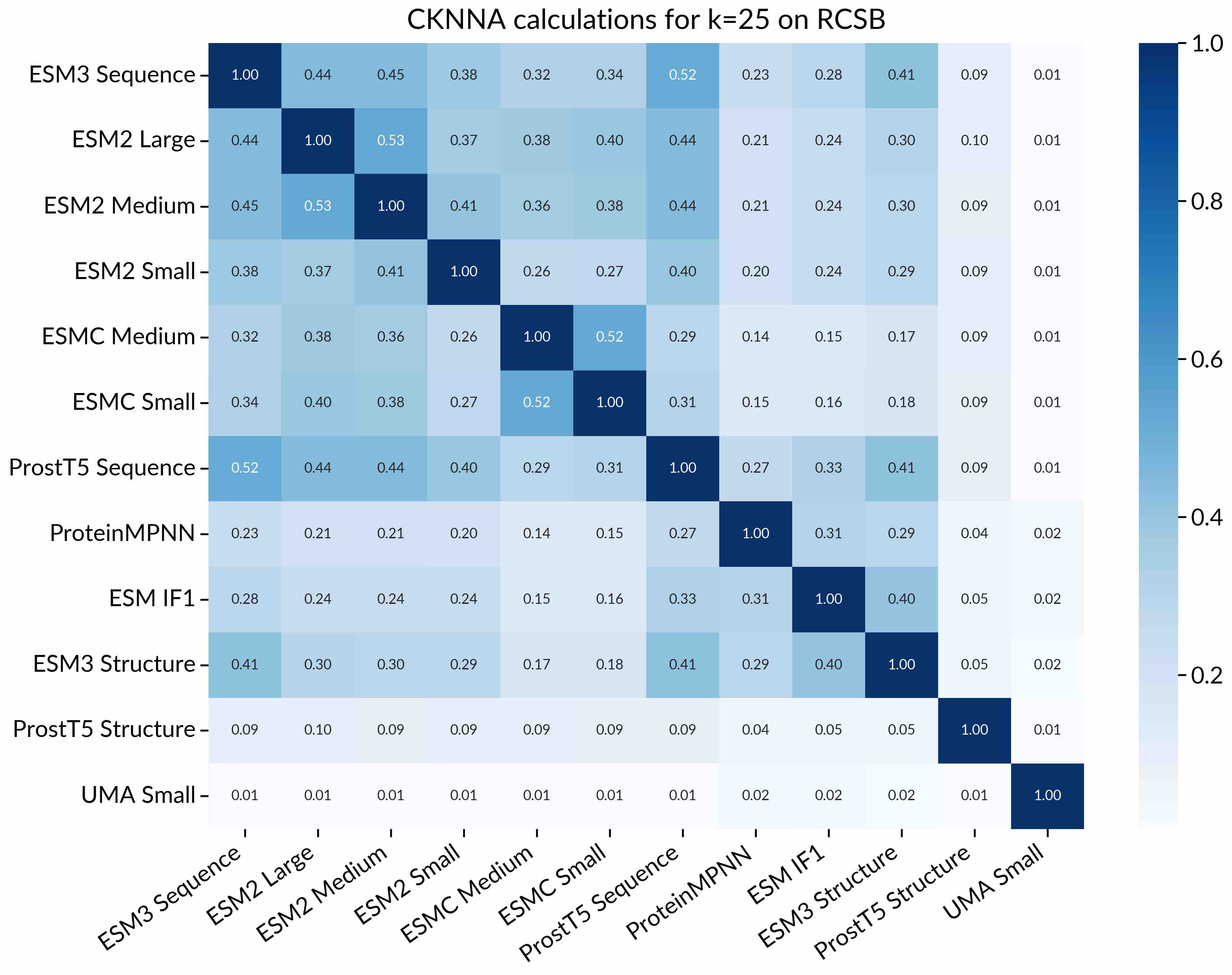}
    \caption{Full CKNNA correlation matrix ($k=25$) between each model's embeddings of the same 50,000 structures randomly sampled from RCSB.}
    \label{fig:RCSB_cknna}
\end{figure}

\end{appendices}



\end{document}